\documentclass[lettersize,journal]{IEEEtran}
\usepackage{amsmath,amsfonts}
\usepackage[linesnumbered,ruled,vlined]{algorithm2e}
\usepackage{array}
\usepackage[caption=false,font=footnotesize,labelfont=sf,textfont=sf]{subfig}
\usepackage{textcomp}
\usepackage{stfloats}
\usepackage{url}
\usepackage{verbatim}
\usepackage{graphicx}
\usepackage{cite}
\usepackage{multirow}
\usepackage{color}
\usepackage{xcolor}
\usepackage{makecell}
\usepackage{gensymb}
\usepackage[switch]{lineno}
\usepackage{booktabs}
\usepackage{dutchcal}
\usepackage{bm}
\newcounter{theorem}
\newenvironment{theorem}{\refstepcounter{theorem}{\noindent\bf Theorem 1 }}
{\hfill $\square$\par}
\definecolor{mycolor}{HTML}{31A5C2}
\begin{document}

\markboth{Submitted to IEEE Transactions on Geoscience and Remote Sensing}%
{Shell \MakeLowercase{\textit{et al.}}: A Sample Article Using IEEEtran.cls for IEEE Journals}

\title{Multisource Semisupervised Adversarial Domain Generalization Network for Cross-Scene Sea\textendash Land Clutter Classification}

\author{Xiaoxuan Zhang,~\IEEEmembership{Student Member,~IEEE},
Quan Pan,~\IEEEmembership{Member,~IEEE},
Salvador Garc\'ia
	
\thanks{This work was supported in part by the China Scholarship Council (CSC) under Grant 202306290161. \textit{(Corresponding author: Xiaoxuan Zhang.)}}

\thanks{Xiaoxuan Zhang and Quan Pan are with the School of Automation and the Key Laboratory of Information Fusion Technology, Ministry of Education, Northwestern Polytechnical University, Xi'an 710072, China~(e-mail: xiaoxuanzhang@mail.nwpu.edu.cn; quanpan@nwpu.edu.cn).}

\thanks{Salvador Garc\'ia is with the Department of Computer Science and
Artificial Intelligence, University of Granada, Granada 18071, Spain (e-mail:
salvagl@decsai.ugr.es).}
}

\maketitle

\begin{abstract}
Deep learning (DL)-based sea\textendash land clutter classification for sky-wave over-the-horizon-radar (OTHR) has become a novel research topic.
In engineering applications, real-time predictions of sea\textendash land clutter with existing distribution discrepancies are crucial.
To solve this problem, this article proposes a novel Multisource Semisupervised Adversarial Domain Generalization Network (MSADGN) for cross-scene sea\textendash land clutter classification.
MSADGN can extract domain-invariant and domain-specific features from one labeled source domain and multiple unlabeled source domains, and then generalize these features to an arbitrary unseen target domain for real-time prediction of sea\textendash land clutter.
Specifically, MSADGN consists of three modules: domain-related pseudolabeling module, domain-invariant module, and domain-specific module.
The first module introduces an improved pseudolabel method called domain-related pseudolabel, which is designed to generate reliable pseudolabels to fully exploit unlabeled source domains.
The second module utilizes a generative adversarial network (GAN) with a multidiscriminator to extract domain-invariant features, to enhance the model's transferability in the target domain.
The third module employs a parallel multiclassifier branch to extract domain-specific features, to enhance the model's discriminability in the target domain.
The effectiveness of our method is validated in twelve domain generalizations (DG) scenarios.
Meanwhile, we selected 10 state-of-the-art DG methods for comparison.
The experimental results demonstrate the superiority of our method.
\end{abstract}


\section{Introduction}
\label{sec: Introduction}
\IEEEPARstart{T}{he} rapid advancement of deep learning (DL) has brought significant opportunities for the development of remote sensing methods \cite{zhu2017deep}.
DL-based sea\textendash land clutter classification for sky-wave over-the-horizon-radar (OTHR), as a branch of remote sensing, has recently emerged as a novel research topic.
The sea\textendash land clutter classification involves identifying whether the background clutter in each range-azimuth cell originates from the sea or land.
Matching these classification results with existing geographic information provides coordinate registration parameters essential for target localization, which holds considerable potential for cost-effectively improving target localization accuracy \cite{guo2021improved}.

\subsection{Related Works on DL-based Sea\textendash Land Clutter Classification}
\label{subsec: Related Works on DL-based SeaLand Clutter Classification}
Over the past five years, remarkable achievements have been made in DL-based sea\textendash land clutter classification.
A large amount of work \cite{li2019sea}, \cite{li2022cross}, \cite{zhang2023data}, \cite{zhang2023triple}, \cite{zhang2024a} has delved into the various challenges encountered in sea\textendash land clutter classification, such as few-shot learning \cite{zhang2023data}, class imbalance learning \cite{zhang2023data}, cross-domain classification \cite{zhang2023triple}, and cross-scale classification \cite{li2022cross}.
DL-based methods include deep convolutional neural networks (DCNNs) \cite{krizhevsky2012imagenet}, generative adversarial networks (GANs) \cite{goodfellow2014generative}, and variational autoencoders (VAEs) \cite{kingma2014auto}.
Li et al. \cite{li2019sea}, \cite{li2022cross} proposed DCNN-based fully supervised learning methods, capable of achieving end-to-end sea\textendash land clutter classification.
The DCNN-based method in \cite{li2019sea} constructed a network with three convolutional layers followed by three fully connected (FC) layers, outperforming traditional sea\textendash land clutter classification methods, including support vector machine (SVM)-based \cite{jin2012svm} and least mean square (LMS)-based \cite{turley2013high} methods.
The DCNN-based method in \cite{li2022cross} implemented cross-scale sea\textendash land clutter classification based on algebraic multigrid within the ResNet-18 framework.
Zhang et al. \cite{zhang2023data}, \cite{zhang2023triple}, \cite{zhang2024a} introduced GAN-based and VAE-based semisupervised and unsupervised learning methods, capable of achieving more challenging sea\textendash land clutter classification.
The GAN-based and VAE-based methods in \cite{zhang2023data} used an auxiliary classifier VAE GAN (AC-VAEGAN) for data augmentation of sea\textendash land clutter, enabling effective few-shot learning and class imbalance learning.
The GAN-based method in \cite{zhang2023triple} used a triple loss adversarial domain adaptation network (TLADAN) to learn domain-invariant features between the source and target domains, successfully achieving cross-domain sea\textendash land clutter classification.
The GAN-based method in \cite{zhang2024a} employed a weighted loss semisupervised GAN (WL-SSGAN) to extract hierarchical features from several unlabeled samples, facilitating the semisupervised classification of sea\textendash land clutter.

\begin{figure*}[!t]
\centering
\includegraphics[width=5in]{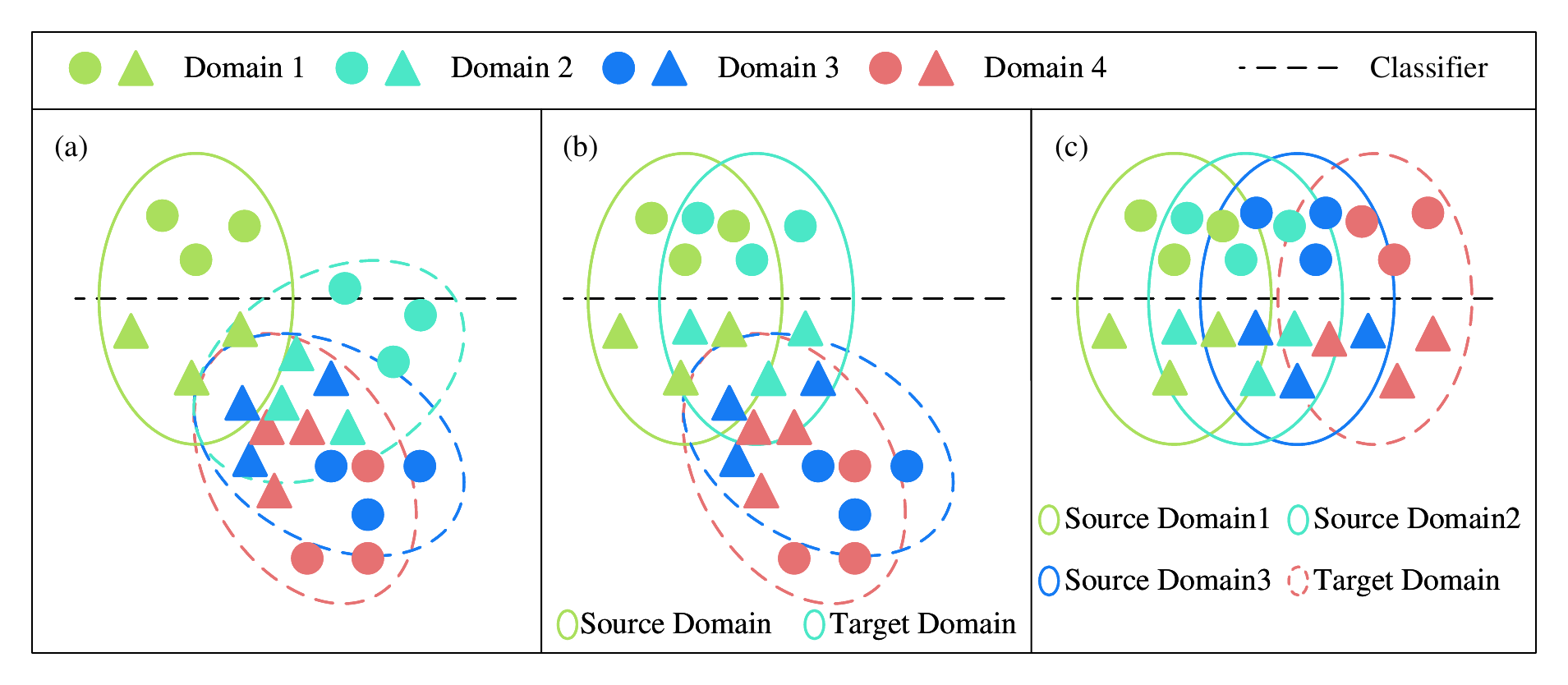}
\caption{Taxonomy of DL-based sea\textendash land clutter classification methods.
(a) ERM-based.
(b) DA-based.
(c) DG-based.
(Best viewed in color.)}
\label{fig: Taxonomy}
\end{figure*}

In summary, the existing DL-based sea\textendash land clutter classification methods can be divided into two categories: 1) empirical risk minimization (ERM)-based methods \cite{li2019sea}, \cite{li2022cross}, \cite{zhang2023data}, \cite{zhang2024a} and 2) domain adaptation (DA)-based methods \cite{zhang2023triple}.
ERM is a common situation in which training and test data follow the independent and identically distributed (i.i.d.) assumption.
However, due to various operating conditions or noise interference, a distribution discrepancy between training and test data inevitably emerges.
In this situation, the performance of the ERM-based method is significantly decreased.
As shown in Fig.~\ref{fig: Taxonomy}(a), the ERM-based method excels at accurately classifying the training set but performs poorly on three different test sets.
To overcome this dilemma, DA-based methods have emerged, that mitigate the distribution discrepancy between the source domain (training set) and the target domain (test set) by learning domain-invariant features.
As shown in Fig.~\ref{fig: Taxonomy}(b), the DA-based method excels at accurately classifying the source domain and performs well on the target domain.

\subsection{Related Works on DA and DG}
\label{subsec: Related Works on DA and DG}
DA-based methods have made significant progress in fields related to sea\textendash land clutter or OTHR, such as computer vision (CV), fault diagnosis, and remote sensing.
In general, according to different strategies for aligning feature distributions, DA-based methods can be divided into three categories \cite{hassanpour2023survey}, \cite{wilson2020survey}: 1) discrepancy-based DA; 2) reconstruction-based DA; and 3) adversarial-based DA.
The discrepancy-based DA aims to reduce the distance or divergence between different distributions by selecting certain metrics, such as maximum mean discrepancy (MMD) \cite{tzeng2014deep}, multiple kernel MMD (MK-MMD) \cite{long2015learning}, deep correlation alignment (Deep CORAL) \cite{sun2016deep}, and central moment discrepancy (CMD) \cite{zellinger2017central}.
For instance, Wan et al. \cite{wan2022novel} proposed an MK-MMD-driven deep convolution multiadversarial DA(DCMADA) network for rolling bearing fault diagnosis.
Yan et al. \cite{yan2019cross} presented a cross-domain distance metric learning (CDDML) framework for scene classification of aerial images.
The reconstruction-based DA strives to learn domain-invariant features by accurately classifying the labeled source domain and reconstructing the source and target domains \cite{ghifary2016deep}.
For instance, Zhu et al. \cite{zhu2023unsupervised} introduced a cycle-consistent adversarial DA network (CCADAN) to reconstruct the source and target domains to achieve intelligent bearing fault diagnosis.
The adversarial-based DA attempts to confuse the features between the source and target domains by employing GAN's adversarial training strategy \cite{ganin2016domain}.
For instance, Wang et al. \cite{wang2020intelligent} used a deep adversarial DA network (DADAN) to transfer fault diagnosis knowledge.
Huang et al. \cite{huang2023cross} presented a spatial-spectral weighted adversarial DA (SSWADA) network for cross-domain wetland mapping.
Zhang et al. \cite{zhang2023triple} proposed a TLADAN for cross-domain sea\textendash land clutter classification and high-resolution remote sensing image classification.

However, DA-based methods are only suitable for offline scenarios, designed for a specific target domain without good generalization capability.
As shown in Fig.~\ref{fig: Taxonomy}(b), DA-based methods exhibit poor performance on two unseen domains.
Domain generalization (DG), a more practical yet challenging topic, involves extracting domain-invariant features from multisource domains and generalizing them to an unseen target domain [see Fig.~\ref{fig: Taxonomy}(c)].
Therefore, DG-based methods can make real-time predictions for unseen target domains.
In the context of sea\textendash land clutter classification, to emphasize the specificity of methods, problems addressed using DA-based methods are termed cross-domain classification, while those addressed using DG-based methods are termed cross-scene classification.
Commonly, DG-based methods can be divided into three categories \cite{wang2023generalizing}: 1) data-based DG; 2) feature-based DG; and 3) learning strategy-based DG.
The data-based DG includes data augmentation and data generation.
For example, Zhou et al. \cite{zhou2021domain} proposed a mixup neural network for cross-scene image classification.
This method can generate new domains in the feature space, thus achieving data augmentation.
Fan et al. \cite{fan2024deep} presented a deep mixed DG network (DMDGN) for intelligent fault diagnosis, in which data augmentation is employed for both class and domain spaces.
The feature-based DG includes domain-invariant representation learning and feature disentanglement.
For example, Zhu et al. \cite{zhu2023style} proposed a style and content separation network (SCSN) for
cross-scene remote sensing image classification.
By effectively disentangling and recognizing the style and content information, SCSN improves the model's generalization capability.
The learning strategy-based DG includes meta-learning, self-supervised learning and ensemble learning.
For example, Ren et al. \cite{ren2024meta} introduced a model-agnostic meta-learning-based training framework called Meta-GENE for fault diagnosis DG.

\subsection{Motivations and Contributions}
\label{subsec: Background}
The most relevant to this article is our latest proposed TLADAN \cite{zhang2023triple}, which considers a distribution discrepancy between training and test data.
As mentioned earlier, TLADAN is only applicable for the offline scenario.
Unfortunately, for OTHR, various working conditions or noise interference may give rise to many unseen target domains.
Thus, a more generalized DG model, as shown in Fig.~\ref{fig: Taxonomy}(c), is desired to address cross-scene sea\textendash land clutter classification.
Furthermore, manual labeling of collected sea\textendash land clutter data is extremely expensive and time-consuming.
A more intricate setting called semisupervised DG (SSDG) is considered.
In this setting, one labeled source domain and multiple unlabeled source domains are available.
Therefore, our motivation is to explore how to utilize these multisource domains to design an SSDG model that works well on an arbitrary unseen target domain.

To this end, this article proposes a feature-based DG method called Multisource Semisupervised Adversarial DG Network (MSADGN), which is capable of extracting domain-invariant and domain-specific features for generalization to the unseen target domain.
Specifically, MSADGN consists of three submodules: domain-related pseudolabeling module, domain-invariant module, and domain-specific module.
In the first submodule, we propose an improved pseudolabel method named domain-related pseudolabel to generate pseudolabels for unlabeled source domains.
The deep semisupervised DG Network (DSDGN) \cite{liao2020deep} utilizes the traditional pseudolabel method to assign pseudolabels for unlabeled source domains.
However, it does not consider distribution differences between these domains; thus, the quality of the generated pseudolabels may be poor.
To solve this problem, we introduce a domain-related similarity score into the traditional pseudolabel method, incorporating a global iteration scheme and a dynamic threshold to generate reliable pseudolabels.
In the second submodule, we integrate an improved domain adversarial neural network (DANN) \cite{ganin2015unsupervised} into MSADGN.
As an adversarial-based DA method, DANN achieves DA by confusing the features between the source and target domains.
Inspired by this, a multidiscriminator is designed to learn domain-invariant features between multisource domains.
This is achieved by learning domain-invariant features between any two domains.
In the third submodule, we integrate a multiclassifier branch to learn domain-specific features, to further enhance the model's generalization capability.
The motivation is that the sea\textendash land clutter from the unseen target domain may share similarities with the domain-specific features of one or more domains in the multisource domains.
We hope to explore more comprehensive features that are both domain-invariant and domain-specific, enabling MSADGN to generalize well to the unseen target domain.

The main contributions of this article can be summarized as follows.

\begin{itemize}
    \item[1)] For the first time, a novel DG architecture, MSADGN, is proposed for cross-scene sea\textendash land clutter classification.
    The proposed MSADGN can extract comprehensive domain-invariant and domain-specific features from one labeled source domain and multiple unlabeled source domains, thereby exhibiting strong transferability and discriminability in the unseen target domain.
    
    \item[2)] Most DG-based methods assume that labels from multisource domains are available.
    This article explores a more complex SSDG scenario.
    Considering the distribution differences of sea\textendash land clutter between different domains, an improved pseudolabel method called domain-related pseudolabel is proposed to generate reliable pseudolabels, thus indirectly obtaining a fully labeled multisource domain for DG.

    \item[3)] Most DG-based methods focus on learning domain-invariant features from multisource domains.
    MSADGN is capable of learning domain-invariant features through a multidiscriminator branch and domain-specific features through a multiclassifier branch.
    Consequently, the real-time prediction results of sea\textendash land clutter can benefit from the learned domain-specific features.
    
    \item[4)] To the best of our knowledge, the cross-scene classification problem of sea\textendash land clutter has not been considered in the literature.
    The effectiveness of MSADGN is verified in twelve DG scenarios using both the 1-D cross-scene sea\textendash land clutter signal (CS-SLCS) dataset and the publicly available 2-D cross-scene high-resolution remote sensing image (CS-HRRSI) dataset.
    Meanwhile, the superiority of MSADGN is validated through the comparison with ten state-of-the-art methods, including five DA-based DG methods and five pure SSDG methods.
\end{itemize}

The rest of this article is organized as follows.
The preliminaries, including problem formulation, GAN, and semisupervised pseudolabel, are presented in Section~\ref{sec: Preliminaries}.
In Section~\ref{sec: Proposed Method}, we elaborate on the proposed MSADGN, including the submodule details, algorithmic procedure, and theoretical insight.
In Section~\ref{sec: Experiments and Analysis},
the effectiveness and superiority of MSADGN are validated using CS-SLCS and CS-HRRSI.
Section~\ref{sec: Conclusion} summarizes this article.

\section{Preliminaries}
\label{sec: Preliminaries}
\subsection{Problem Formulation}
\label{subsec: Problem Formulation}
DG is more challenging than DA in the cross-scene sea\textendash land clutter classification task, because there is no prior knowledge of the target domain.
DG extracts domain-invariant (and domain-specific) features from multisource domains ${\mathcal D}_s = \{{\mathcal D}_s^k\}_{k=1}^K$ and generalize them to an arbitrary unseen target domain ${\mathcal D}_t$, where $K$ is the number of multisource domains.
This article focuses on an SSDG problem, where only the labels of ${\mathcal D}_s^1$ are accessible, whereas the labels of the other $\{{\mathcal D}_s^k\}_{k=2}^K$ are inaccessible.
In other words, ${\mathcal D}_s^1=\{({\mathbf x}_i^l,y_i^l,d_i^l=1)\}_{i=1}^{n_l}$ and ${\mathcal D}_s^{k\neq 1}=\{({\mathbf x}_i^u,d_i^u=k)\}_{i=1}^{n_k}$, where ${\mathbf x}_i^*$, $y_i^l$, $d_i^*$, $n_l$ and $n_k$ denote an input sample, a category label, a domain label, the number of labeled samples in ${\mathcal D}_s^1$, and number of unlabeled samples in ${\mathcal D}_s^{k\neq 1}$, respectively.
For different domains, the label space remains the same, whereas the feature space or data distribution tends to be different, i.e. $P_s^i({{\mathbf X}})\neq P_s^j({{\mathbf X}})\neq P_t({{\mathbf X}}), 1\le i\neq j\le K$, where $P_s^*({\mathbf X})$ denotes the source distribution and $P_t({\mathbf X})$ denotes the target distribution.
The goal of this article is to develop a generalizable MSADGN to minimize the risk of unseen target classification, $\mathbb E_{({\mathbf x},y)\in {\mathcal D}_t}[\mathcal F({\mathbf x})\neq y]$, where $\mathbb E$ represents the expectation operator.
Table~\ref{table: Notations} lists the most commonly used notations in this article along with their explanations.

\begin{table*}[!t]
\setlength{\tabcolsep}{3pt}
\centering
\caption{The Most Commonly Used Notations and Their Explanations}
\label{table: Notations}
\renewcommand\arraystretch{1.3}
\begin{tabular}{|ll|}
\hline
\makebox[0.12\textwidth][l]{\bf Notations} & \makebox[0.5\textwidth][l]{\bf Explanations}\\
\hline
${\mathcal D}_s$ & Multisource domains\\
${\mathcal D}_s^1$ & A labeled source domain in ${\mathcal D}_s$\\
${\mathcal D}_s^{k\neq 1}$ & Unlabeled source domains in ${\mathcal D}_s$\\
${\mathcal D}_{s'}$ & Multisource domains output by the domain-related pseudolabeling module\\
${\mathcal D}_{s'}^1$ & A labeled source domain in ${\mathcal D}_{s'}$\\
${\mathcal D}_{s'}^{k\neq 1}$ & Pseudolabeled source domains in ${\mathcal D}_{s'}$\\
${\mathcal D}_t$ & An unseen target domain\\
$({\mathbf x}_i^l,y_i^l,d_i^l=1)$ & A labeled sample in ${\mathcal D}_s^1$, where ${\mathbf x}_i^l$, $y_i^l$ and $d_i^l$ denote labeled sample, category label and domain label\\
$({\mathbf x}_i^u,d_i^u=k)$ & An unlabeled sample in ${\mathcal D}_s^{k\neq 1}$, where ${\mathbf x}_i^u$ and $d_i^u$ denote unlabeled sample and domain label\\
$({\mathbf x}_i^u,\widehat{y_i^u},d_i^u=k)$ & A pseudolabeled sample in ${\mathcal D}_{s'}^{k\neq 1}$, where ${\mathbf x}_i^u$, $\widehat{y_i^u}$ and $d_i^u$ denote pseudolabeled sample, pseudolabel and domain label\\
$n_l$ & The number of labeled samples in ${\mathcal D}_s^1$\\
$n_k$ & The number of unlabeled samples in ${\mathcal D}_s^{k\neq 1}$\\
$n'_k$ & The number of pseudolabeled samples in ${\mathcal D}_{s'}^{k\neq 1}$\\
${\mathbf f}_k$ & A domain-related feature embedding\\
${\mathbf M}_1$ & A category-related feature embedding\\
${\mathbf z}_{i,k}$ & A domain-specific feature for ${\mathbf x}_i\in {\mathcal D}_{s'}$ in domain $k$\\
${\mathbf z}_i$ & A weighted domain-specific feature for ${\mathbf x}_i\in {\mathcal D}_{s'}$\\
${\mathbf w}_i$ & A domain-related similarity weight for ${\mathbf x}_i\in {\mathcal D}_{s'}$\\
$K$ & The number of multisource domains\\
$C$ & The number of classes in each domain\\
$L$ & The dimension of feature embedding in ${\mathcal D}_s^1$\\
$I$ & The number of training iterations\\
$\phi({\mathbf x}_i^u)$ & A probability score in the domain-related pseudolabeling module\\
$\psi({\mathbf x}_i^u)$ & A similarity score in the domain-related pseudolabeling module\\
$\Phi({\mathbf x}_i^u)$ & A pseudolabel score in the domain-related pseudolabeling module\\
$\tau_0$ & A fixed threshold in the traditional pseudolabel method\\
$\tau$ & A dynamic threshold in the domain-related pseudolabeling module\\
${\mathcal l}_{\text{CE}}$ & The crossentropy loss\\
${\mathcal L}_{\text{inv}}$ & The domain-invariant loss of MSADGN\\
${\mathcal L}_{\text{spe}}$ & The domain-specific loss of MSADGN\\
$\mathcal L$ & The overall loss of MSADGN\\
$\alpha$ & The tradeoff parameter between $\phi({\mathbf x}_i^u)$ and $\psi({\mathbf x}_i^u)$\\
$\lambda$ & The tradeoff parameter between ${\mathcal L}_{\text{inv}}$ (or ${\mathcal L}_{\text{adv}}$) and ${\mathcal L}_{\text{spe}}$\\
$\rho_I$ & The tradeoff parameter between ${\mathbf M}_1^{(I)}$ and ${\mathbf M}_1^{(I-1)}$\\
$F_{\text{shared}}$, $F_{\text{weighted}}$ & A shared feature extractor and a weighted feature extractor\\
$D_1$, $D_2$, $\dots$, $D_{K(K-1)/2}$ & Domain classifiers\\
$C_1$, $C_2$, $\dots$, $C_K$ & Task-specific classifiers\\
$C_{\text{weighted}}$ & A weighted classifier\\
$\theta_{f_1}$, $\theta_{f_2}$ & The parameters of $F_{\text{shared}}$ and $F_{\text{weighted}}$\\
$\theta_{d_1}$, $\theta_{d_2}$, $\dots$, $\theta_{d_{K(K-1)/2}}$ & The parameters of $D_1$, $D_2$, $\dots$, and $D_{K(K-1)/2}$\\
$\theta_{c_1}$, $\theta_{c_2}$, \dots, $\theta_{c_K}$ & The parameters of $C_1$, $C_2$, \dots, and $C_K$\\
$\theta_{c_{K+1}}$ & The parameter of $C_{\text{weighted}}$\\
\hline
\end{tabular}
\end{table*}

\subsection{GAN}
\label{subsec: GAN}
The original GAN \cite{goodfellow2014generative} consists of two adversarial neural networks: a generator ($G$) and a discriminator ($D$).
$G$ generates realistic samples, whereas $D$ strives to differentiate between samples generated by $G$ and real samples.
In this way, adversarial training of the two networks can be described as a two-player minimax game with the following value function:
\begin{equation}
\label{equ: GAN}
\begin{aligned}
\min\limits_G \max\limits_D V(D, G) = &{\mathbb E_{{\mathbf x}\sim p_{_{\rm data}}}[\log D({\mathbf x})]} \\
&+ {\mathbb E_{{\mathbf z}\sim p_{_{\mathbf z}}}[1- \log D(G({\mathbf z}))]}
\end{aligned}
\end{equation}
where $p_{_{\rm data}}$ is the distribution of real data ${\mathbf x}$, $p_{_{\mathbf z}}$ is the distribution of prior noise ${\mathbf z}$, $D({\mathbf x})$ denotes the probability that ${\mathbf x}$ comes from the real data, $G({\mathbf z})$ denotes the data generated by $G$,
and $D(G({\mathbf z}))$ denotes the probability that $G({\mathbf z})$ comes from the generated data.

In the MSADGN submodule, GAN is employed to extract domain-invariant features.
Specifically, $G$ is used to extract features from the multisource domains.
In contrast, $D$ distinguishes from which source domain these features originate.

\subsection{Semisupervised Pseudolabel}
\label{subsec: Semisupervised Pseudolabel}
Pseudolabel \cite{lee2013pseudo} is a prevalent semisupervised learning method that utilizes unlabeled samples to enhance the performance of deep neural networks (DNNs).
The core idea behind this method is to use existing models to predict unlabeled samples, treat these predictions as pseudolabels, and then merge them with labeled samples for further model training.
Specifically, the traditional pseudolabel selects the token with the maximum prediction probability as the pseudolabel for each unlabeled sample. The one-hot coding pseudolabel $\widehat{{\mathbf y}^u}=[\widehat{y_1^u}, \widehat{y_2^u}, \dots, \widehat{y_C^u}]$ can be expressed as follows:
\begin{equation}
\label{equ: pseudolabel}
\widehat{y_j^u}=\begin{cases}
    1 & \text{if}~j=\text{argmax}_{j'}\phi_{j'}({\mathbf x}^u)~\text{and}~\phi_{j'}({\mathbf x}^u)>\tau_0 \\
    0 & \text{otherwise}~~j=1, 2, \dots, C
\end{cases}
\end{equation}
where ${\mathbf x}^u$ is an input unlabeled sample, $\phi_{j'}({\mathbf x}^u)$ denotes the $j'$-th element of the output vector $\phi({\mathbf x}^u)$, $\tau_0$ is a fixed threshold, and $C$ is the number of classes.

\section{Proposed MSADGN for Cross-Scene Sea\textendash Land Clutter Classification}
\label{sec: Proposed Method}
In this section, we first introduce the overall architecture of MSADGN, followed by a detailed description of its submodules, including the domain-related pseudolabeling module, domain-invariant module, and domain-specific module.
Next, the training procedure of MSADGN is presented.
Finally, we provide a theoretical insight into MSADGN.

\subsection{Overall Architecture of MSADGN}
\label{subsec: Overall Architecture of MSADGN}
\begin{figure*}[!t]
\centering
\includegraphics[width=7in]{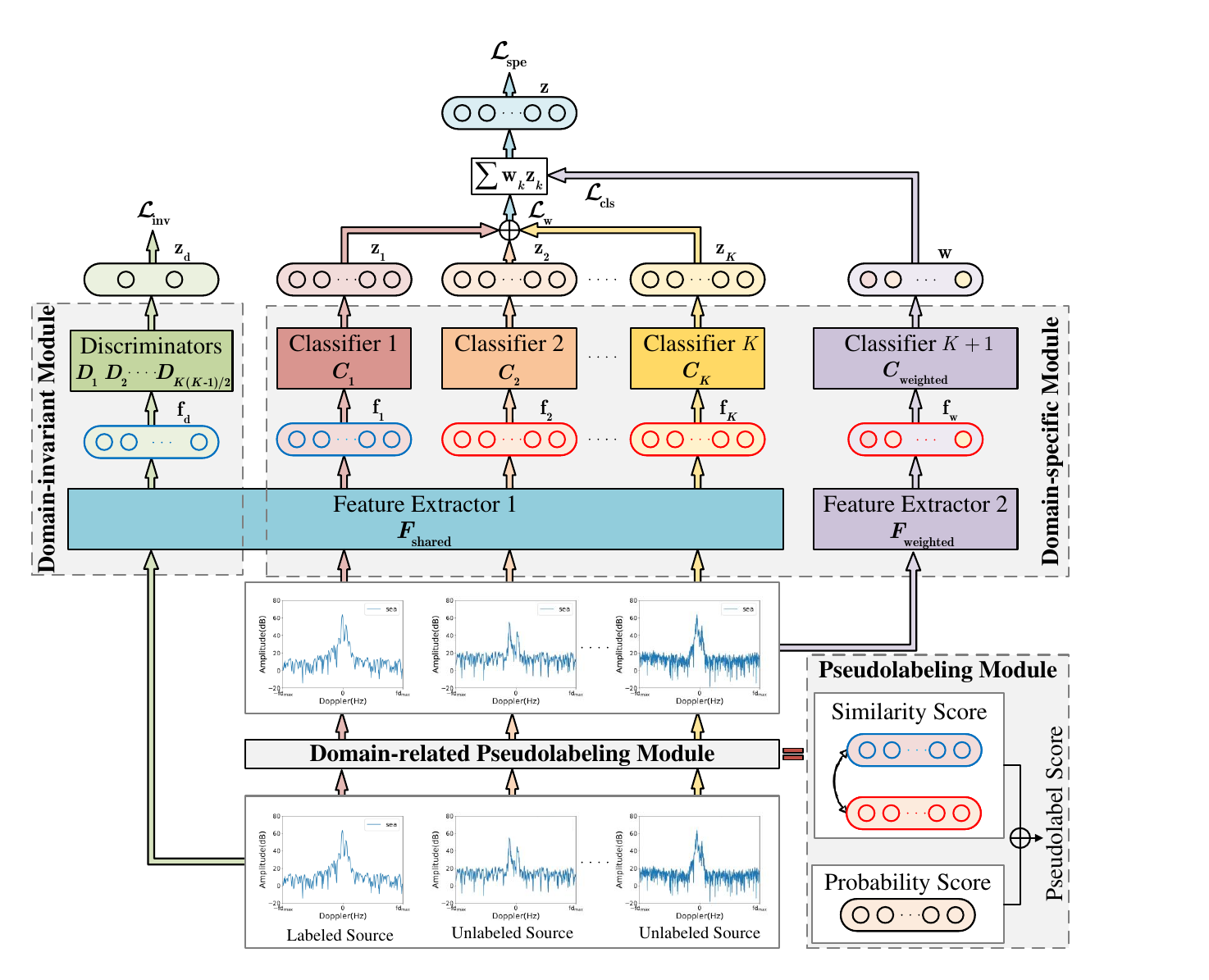}
\caption{Illustration of cross-scene sea\textendash land clutter classification via the proposed MSADGN. (Best viewed in color.)}
\label{fig: Proposed MSADGN}
\end{figure*}

The overall architecture of MSADGN for cross-scene sea\textendash land clutter classification is illustrated in Fig.~\ref{fig: Proposed MSADGN}, which consists of three modules: domain-related pseudolabeling module, domain-invariant, and domain-specific modules.
In more detail, MSADGN comprises two feature extractors, $K+1$ classifiers, and $K(K-1)/2$ discriminators.
Feature extractor 1 is the shared feature extractor $F_{\text{shared}}$, feature extractor 2 is the weighted feature extractor $F_{\text{weighted}}$, the first $K$ classifiers are task-specific classifiers $C_1$, $C_2$, $\dots$, and $C_K$, the ($K+1$)-th classifier is the weighted classifier $C_{\text{weighted}}$, and the $K(K-1)/2$ discriminators represent the domain classifiers $D_1$, $D_2$, $\dots$, and $D_{K(K-1)/2}$.

The input sea\textendash land clutter samples consist of two parts: labeled samples from ${\mathcal D}_s^1$ and unlabeled samples from ${\mathcal D}_s^{k\neq 1}$.
On the one hand, the input samples enter the domain-related pseudolabeling module, where unlabeled samples with pseudolabel scores exceeding a certain threshold are selected and assigned pseudolabels.
As a result, labeled samples from all source domains are indirectly acquired, and we denote the current multisource domains as ${\mathcal D}_{s'} = \{{\mathcal D}_{s'}^k\}_{k=1}^K$.
Conversely, since the domain-invariant module only requires samples with domain labels, we take all samples from ${\mathcal D}_s$ as input into this module to extract domain-invariant features.
Furthermore, to enhance the model's generalization capability, the domain-specific module is integrated, and all labeled samples from
${\mathcal D}_{s'}$ are input into this module to extract domain-specific features.
Because of the more comprehensive features extracted by the domain-invariant and domain-specific modules, our model exhibits better transferability and discriminability in the unseen target domain.

\subsection{Domain-Related Pseudolabeling Module}
\label{subsec: Domain-Related Pseudolabeling Module}
The domain-related pseudolabeling module comprises $F_{\text{shared}}$ and $C_1$.
The traditional pseudolabel method is applicable when the training samples follow the i.i.d. assumption.
However, in the DG problem, due to various working conditions or noise interference, training samples from multisource domains may have different probability distributions.
If pseudolabels are assigned fairly to unlabeled samples in ${\mathcal D}_s^{k\neq 1}$ using the aforementioned method without considering domain gaps, poor pseudolabels may be obtained.
To solve this problem, we propose an improved pseudolabel method called domain-related pseudolabel.
In the proposed method, if the pseudolabel score $\Phi({\mathbf x}_i^u)$ exceeds a certain threshold, pseudolabels are assigned unfairly to unlabeled samples in ${\mathcal D}_s^{k\neq 1}$ as we expect, where ${\mathbf x}_i^u\in {\mathcal D}_s^{k\neq 1}$.
To this end, we need to gradually adjust the values of the following three components during the training process: 1) probability score $\phi({\mathbf x}_i^u)$; 2) similarity score $\psi({\mathbf x}_i^u)$; and 3) dynamic threshold $\tau$.

\subsubsection{Probability Score}
\label{subsubsec: Probability Score}
We still obtain $\phi({\mathbf x}_i^u)$ using the traditional pseudolabel method.
In the domain-related pseudolabeling module, the initial step involves updating both $F_{\text{shared}}$ and $C_1$ using labeled samples from $\mathcal D_s^1$.
Subsequently, $\phi({\mathbf x}_i^u)$ can be obtained from $\phi({\mathbf x}_i^u) = softmax(C_1(F_{\text{shared}}({\mathbf x}_i^u)))$, where $softmax(\cdot)$ represents the softmax activation function.

\subsubsection{Similarity Score}
\label{subsubsec: Similarity Score}
Further, $\psi({\mathbf x}_i^u)$ is introduced as a correction term for $\phi({\mathbf x}_i^u)$ to consider domain-related information.
Our motivation is that the domain-related feature embedding ${\mathbf f}_k = F_\text{shared}({\mathbf x}_i^u)\in{\mathbb R}^{1\times L}$ in ${\mathcal D}_s^{k\neq 1}$ is supposed to gather around the category-related feature embeddings ${\mathbf M}_1\in{\mathbb R}^{L\times C}$ in ${\mathcal D}_s^1$, where $L$ represents the feature embedding dimensions.
And ${\mathbf M}_1$ is defined as follows:
\begin{equation}
\label{equ: M_1}
\begin{aligned}
&{\mathbf M}_1 = [{\mathbf f}_1(c=1), {\mathbf f}_1(c=2), \dots, {\mathbf f}_1(c=C)]^\text{T}\\
&{\mathbf f}_1(c=j) = \frac{1}{n_l^j}\sum\nolimits_{({\mathbf x}_i^l, y_i^l=j)\in {\mathcal D}_s^1}F_{\text{shared}}({\mathbf x}_i^l)
\end{aligned}
\end{equation}
where $n_l^j$ denotes the number of labeled samples in the $j$-th class.
Then, $\psi({\mathbf x}_i^u)$ can then be obtained by measuring the similarity between ${\mathbf f}_k$ and ${\mathbf M}_1$:
\begin{equation}
\label{equ: psi}
    \psi({\mathbf x}_i^u) = CS({\mathbf f}_k, {\mathbf M}_1), k=2, 3, \dots, K
\end{equation}
where $CS(\cdot, \cdot)$ denotes the cosine similarity function.

A straightforward way to compute $\psi({\mathbf x}_i^u)$ is by measuring the similarity between ${\mathbf f}_k$ and ${\mathbf M}_1$ obtained locally from the current mini-batch of labeled samples from ${\mathcal D}_s^1$.
However, the limited number of samples in the mini-batch may result in an inadequate ${\mathbf M}_1$. For example, the current mini-batch might not include feature embeddings for certain categories.
In addition, even one labeled sample with high-confidence misclassification may cause high similarity bias.

To address this problem, we propose a global iteration scheme to obtain ${\mathbf M}_1$.
First, we reformulate ${\mathbf M}_1$ into a form associated with the number of training iterations $I$:
\begin{equation}
\label{equ: M_1I}
\begin{aligned}
&{\mathbf M}_1^{(I)} = [{\mathbf f}_1^{(I)}(c=1), {\mathbf f}_1^{(I)}(c=2), \dots, {\mathbf f}_1^{(I)}(c=C)]^\text{T}\\
&{\mathbf f}_1^{(I)}(c=j) = \frac{1}{n_l^j}\sum\nolimits_{({\mathbf x}_i^l, y_i^l=j)\in {\mathcal D}_s^1(I)}F_{\text{shared}}({\mathbf x}_i^l)
\end{aligned}
\end{equation}
where ${\mathcal D}_s^1(I)$ represents the mini-batch of ${\mathcal D}_s^1$ in the $I$-th iteration. Then, ${\mathbf M}_1$ is initialized with ${\mathbf M}_1^{(0)}$ and subsequently updated based on the last iteration as follows:
\begin{equation}
\label{equ: M_1g}
\begin{aligned}
    \rho_I &= CS({\mathbf M}_1^{(I)},{\mathbf M}_1^{(I-1)})\\
    {\mathbf M}_1^{(I)} &\leftarrow \rho_I^2 {\mathbf M}_1^{(I)} + (1-\rho_I^2) {\mathbf M}_1^{(I-1)}
\end{aligned}
\end{equation}
where $\rho_I$ is the tradeoff parameter.
The advantage of the global iteration scheme is that if ${\mathbf M}_1^{(I)}$ is inadequate, intuitively, its similarity to ${\mathbf M}_1^{(I-1)}$ should be low, indicating that $\rho_I$ tends to 0.
At this time, according to Eq.~\eqref{equ: M_1g}, one can take ${\mathbf M}_1^{(I-1)}$ as the ${\mathbf M}_1^{(I)}$ in the $I$-th iteration.

\subsubsection{Dynamic Threshold}
\label{subsubsec: Dynamic Threshold}
A suitable threshold is crucial for selecting a reliable pseudolabel.
If the fixed threshold $\tau_0$ mentioned earlier is directly applied, more unlabeled samples with high-confidence misclassification may be selected as the training progresses.
However, a dynamic threshold $\tau$ gradually increasing with training epochs may alleviate the aforementioned potential errors:
\begin{equation}
\label{equ: tau}
\tau = \frac{1}{1 + e^{-10 * p}} - 0.1
\end{equation}
where $p$ is the training progress that changes from 0 to 1.
Here, we set $p=(n_{e-1}*L_e+s_e)/(N_e*L_e)$, where $n_{e-1}$ denotes the last training epoch, $L_e$ denotes the number of data batches in each epoch, $s_e$ denotes the current training step within the current epoch, and $N_e$ is the maximum number of training epochs.

Finally, with the definitions of $\phi({\mathbf x}_i^u)$, $\psi({\mathbf x}_i^u)$ and $\tau$, the domain-related pseudolabel can be selected.
First, the new pseudolabel score $\Phi({\mathbf x}_i^u)$ can be given by the following formula:
\begin{equation}
\label{equ: Phi}
    \Phi({\mathbf x}_i^u) = \alpha\phi({\mathbf x}_i^u) + (1-\alpha)\psi({\mathbf x}_i^u)
\end{equation}
where $\alpha$ is the tradeoff parameter.

Next, the domain-related pseudolabel can be selected in a manner similar to Eq.~\eqref{equ: pseudolabel}. One-hot coding domain-related pseudolabel $\widehat{{\mathbf y}_i^u}=[\widehat{y_{i,1}^u}, \widehat{y_{i,2}^u}, \dots, \widehat{y_{i,C}^u}]$ of $\widehat{y_i^u}$ can be expressed as follows:
\begin{equation}
\label{equ: domain-related pseudolabel}
\widehat{y_{i,j}^u}=\begin{cases}
    1 & \text{if}~j=\text{argmax}_{j'}\Phi_{j'}({\mathbf x}_i^u)~\text{and}~\Phi_{j'}({\mathbf x}_i^u)>\tau \\
    0 & \text{otherwise}~~j=1, 2, \dots, C
\end{cases}
\end{equation}
where $\Phi_{j'}({\mathbf x}_i^u)$ denotes the $j'$-th element of $\Phi({\mathbf x}_i^u)$.

After the domain-related pseudolabeling module, we can obtain the fully labeled multisource domains ${\mathcal D}_{s'} = \{{\mathcal D}_{s'}^k\}_{k=1}^K$, where ${\mathcal D}_{s'}^1={\mathcal D}_s^1$ and ${\mathcal D}_{s'}^{k\neq 1}=\{({\mathbf x}_i^u,\widehat{y_i^u},d_i^u=k)\}_{i=1}^{{n'_k}}$, with $n'_k$ denoting the number of pseudolabeled samples in ${\mathcal D}_{s'}^{k\neq 1}$, which provides sufficient labeled training samples for the subsequent domain-specific module.

\subsection{Domain-Invariant Module}
\label{subsec: Domain-Invariant Module}
The domain-invariant module comprises $F_{\text{shared}}$, $D_1$, $D_2$, $\dots$, and $D_{K(K-1)/2}$.
The design of this module is inspired by DANN.
It is worth noting that DANN is primarily employed for DA, achieving the alignment of feature distributions between the source and target domains through adversarial training.
Since ${\mathcal D}_t$ is unseen during the training phase of MSADGN, one can equivalently treat ${\mathcal D}_s^{k_1}$ as the source domain of DANN, and ${\mathcal D}_s^{k_2}$ as the target domain of DANN, where $1\leq k_1\neq k_2\leq K$.
In other words, $K(K-1)/2$ discriminators are designed to distinguish the features of any two source domains.

In this module, the domain-invariant feature is extracted through adversarial training similar to Eq.~\eqref{equ: GAN} as follows:
\begin{equation}
\label{equ: L_adv}
\begin{aligned}
    {\mathcal L}_{\text{adv}} &= \frac{1}{K(K-1)/2}\sum_{k=1}^{K(K-1)/2}{\mathcal L}_{\text{adv}}^k\\
    {\mathcal L}_{\text{adv}}^k &= \min_{F_{\text{shared}}}\max_{D_k}-\frac{\lambda}{n}\sum_{{\mathbf x}_i\in {\mathcal D}}{\mathcal l}_{\text{CE}}(D_k(F_{\text{shared}}({\mathbf x}_i)), s_i)
\end{aligned}
\end{equation}
where $\lambda$ denotes the tradeoff between ${\mathcal L}_{\text{adv}}$ and the domain-specific loss ${\mathcal L}_{\text{spe}}$ defined in Section \ref{subsec: Domain-Specific Module}, $n=n_{k_1}+n_{k_2}$ represents the total number of samples in ${\mathcal D}={\mathcal D}_s^{k_1}\cup {\mathcal D}_s^{k_2}$,
${\mathcal l}_{\text{CE}}(\cdot,\cdot)$ denotes the crossentropy loss, and $s_i$ represents a binary label of the ${\mathcal D}_s^{k_1}$ and ${\mathcal D}_s^{k_2}$ such that $s_i=1$ if ${\mathbf x}_i\in {\mathcal D}_s^{k_1}$ or $s_i=0$ if ${\mathbf x}_i\in {\mathcal D}_s^{k_2}$.
Here, we set $\lambda = 2 / (1 + \exp(-10*p)) - 1$, as there is a significant distribution gap among multisource domains in the early stages of the training process. The gradually increasing tradeoff $\lambda$ helps prevent the model from overfitting to the labeled source data.

Furthermore, a gradient reversal layer is incorporated into the backpropagation process to facilitate a one-step update, which can be defined using the following pseudofunction:
\begin{equation}
\label{equ: GRL}
\begin{aligned}
    F_{\text{shared}}^\lambda({\mathbf x}) &= {\mathbf x} \\
    \frac{{dF_{\text{shared}}^\lambda({\mathbf x})}}{{d{\mathbf x}}} &=  - \lambda {\bf{I}}
\end{aligned}
\end{equation}
where $\bf{I}$ is an identity matrix.
Then $F_{\text{shared}}$, $D_1$, $D_2$, $\dots$, and $D_{K(K-1)/2}$ can be simultaneously optimized by minimizing the following domain-invariant loss ${\mathcal L}_{\text{inv}}$:
\begin{equation}
\label{equ: L_{inv}}
{\mathcal L}_{\text{inv}} \hspace{-1pt}=\hspace{-1pt} \frac{1}{K(K-1)/2}\sum_{k=1}^{K(K-1)/2}\hspace{-1pt}\frac{1}{n}\hspace{-1pt}\sum_{{\mathbf x}_i\in {\mathcal D}}{\mathcal l}_{\text{CE}}(D_k(F_{\text{shared}}^\lambda({\mathbf x}_i)), s_i)
\end{equation}

\subsection{Domain-Specific Module}
\label{subsec: Domain-Specific Module}
\begin{figure}[!t]
\centering
\includegraphics[width=3.4in]{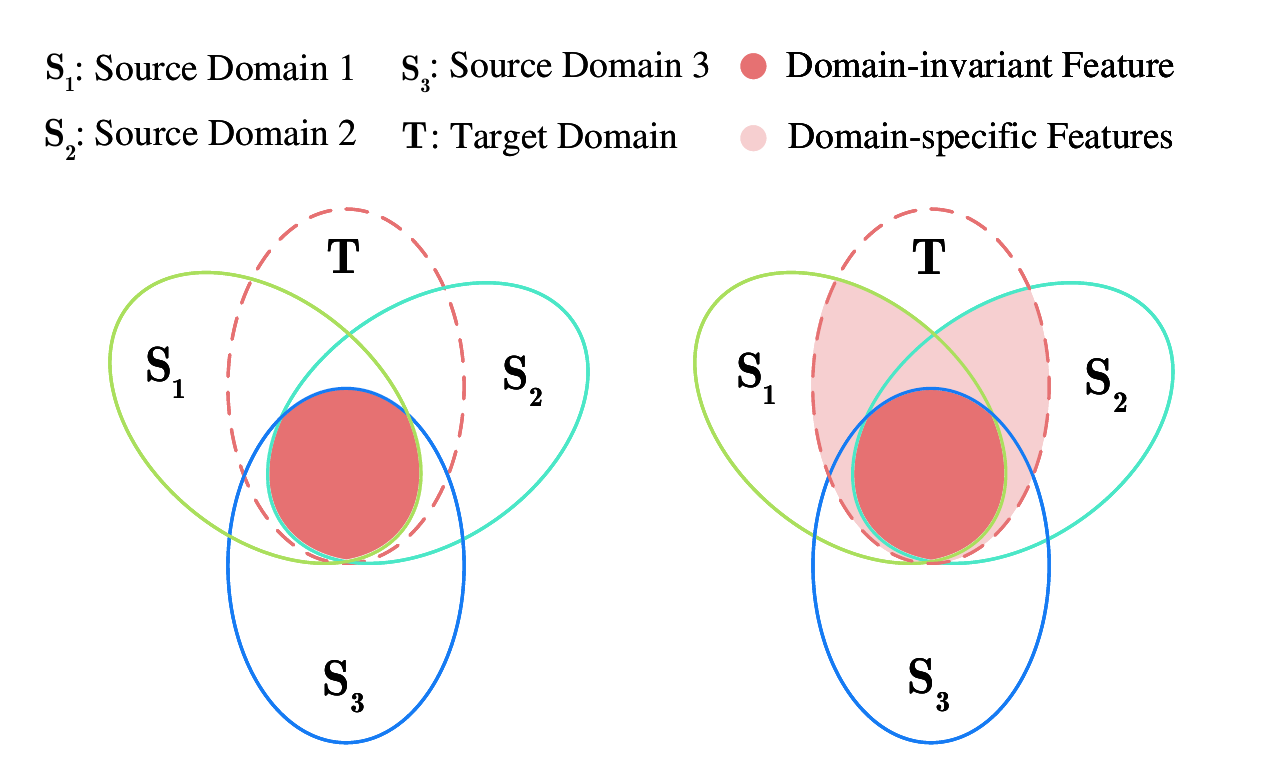}
\caption{Schematic diagram of domain-invariant feature and domain-specific features. (Best viewed in color.)}
\label{fig: inv-spe}
\end{figure}

The domain-specific module comprises $F_{\text{shared}}$, $C_1$, $C_2$, $\dots$, $C_K$, $F_{\text{weighted}}$ and $C_{\text{weighted}}$.
Although we have extracted domain-invariant features through the domain-invariant module, in practical applications, the unseen target domain may share similarities with the domain-specific features of one or more domains in the multisource domains (see Fig.~\ref{fig: inv-spe}).
Thus, we designed a domain-specific module aimed at extracting domain-specific features to further enhance the model's generalization capability.

Specifically, $F_{\text{shared}}$ and $C_k$ are introduced to extract the domain-specific feature ${\mathbf z}_{i,k}=C_k(F_{\text{shared}}({\mathbf x}_i))$, where ${\mathbf x}_i\in {\mathcal D}_{s'}$ and $k=1, 2, \dots, K$.
The features are combined through a domain-related similarity weight ${\mathbf w}_i=\left[w_{i,1}, w_{i,2}, \dots, w_{i, K} \right]^\text{T}$:
\begin{equation}
\label{equ: z}
    {\mathbf z}_i = \sum_{k=1}^K w_{i,k} \cdot {\mathbf z}_{i,k},~ 0\leq w_{i,k}\leq 1~\text{and}~\sum_{k=1}^K w_{i,k}=1
\end{equation}
where $w_{i,k}$ denotes the similarity between the input ${\mathbf x}_i$ and the $k$-th domain.
Next, we design an appropriate ${\mathbf w}_i$ for ${\mathcal D}_{s'}$ and ${\mathcal D}_t$, respectively.

On the one hand, if ${\mathbf x}_i\in {\mathcal D}_{s'}^k$, then its domain label $d_i=k$.
In other words, we can consider that ${\mathbf x}_i$ is exclusively similar to domain $k$ and unrelated to other source domains.
Thus, $w_{i,k}$ can be defined as follows:
\begin{equation}
\label{w_{i,k}}
w_{i,k} = 
    \begin{cases}
    1 & \text{if}~d_i=k\\
    0 & \text{otherwise}~~d_i=1, 2, \dots, K
    \end{cases}
\end{equation}

As a result, the parameters of $F_{\text{shared}}$ and $C_k$ can be updated solely by averaging the losses computed from the labeled samples in ${\mathcal D}_{s'}^k$:
\begin{equation}
\label{equ: L_{cls}}
    {\mathcal L}_{\text{cls}} = \frac{1}{K}\sum_{k=1}^K \frac{1}{n'_k}\sum_{{\mathbf x}_i\in {\mathcal D}_{s'}^k}{\mathcal l}_{\text{CE}}({\mathbf z}_i, y_i)
\end{equation}
where $n'_1=n_l$, and if $({\mathbf x}_i,y_i)\in {\mathcal D}_{s'}^1$ then $({\mathbf x}_i,y_i)=({\mathbf x}_i^l,y_i^l)$; if $({\mathbf x}_i,y_i)\in {\mathcal D}_{s'}^{k\neq 1}$ then $({\mathbf x}_i,y_i)=({\mathbf x}_i^u,\widehat{y_i^u})$.

On the other hand, if ${\mathbf x}_i\in {\mathcal D}_{t}$ and its domain label $d_i=k$, then one can obtain the corresponding category label through the trained $F_{\text{shared}}$ and $C_k$.
Unfortunately, the domain label is unknown at test time.
To solve this problem, a parallel domain classifier, which is composed of $F_{\text{weighted}}$ and $C_{\text{weighted}}$, is incorporated into our deep architecture to learn ${\mathbf w}_i$ between the target domain and the multisource domains as follows:
\begin{equation}
\label{equ: w_i}
    {\mathbf w}_i=C_{\text{weighted}}(F_{\text{weighted}}({\mathbf x}_i))
\end{equation}
where $w_{i,k}$ denotes the probability that the input ${\mathbf x}_i$ belongs to the $k$-th domain, which implies the similarity between ${\mathbf x}_i$ and the $k$-th domain.

Since the domain label $d_i$ for each ${\mathbf x}_i\in {\mathcal D}_{s'}$ is known at the train time, the parameters of $F_{\text{weighted}}$ and $C_{\text{weighted}}$ can be updated by the following loss:
\begin{equation}
\label{equ: L_w}
    {\mathcal L}_\text{w} = \frac{1}{\sum\nolimits_{k=1}^{K}n'_k}\sum_{{\mathbf x}_i\in {\mathcal D}_{s'}}{\mathcal l}_{\text{CE}}({\mathbf w}_i, d_i)
\end{equation}
where if $({\mathbf x}_i,d_i)\in {\mathcal D}_{s'}^1$ then $({\mathbf x}_i,d_i)=({\mathbf x}_i^l,d_i^l=1)$; if $({\mathbf x}_i,d_i)\in {\mathcal D}_{s'}^{k\neq 1}$ then $({\mathbf x}_i,d_i)=({\mathbf x}_i^u,d_i^u=k)$.

Finally, the total domain-specific loss ${\mathcal L}_{\text{spe}}$ can be expressed as follows:
\begin{equation}
\label{equ: L_spe}
    {\mathcal L}_{\text{spe}} = {\mathcal L}_{\text{cls}} + {\mathcal L}_\text{w}
\end{equation}

\subsection{Training Procedure of MSADGN}
\label{subsec: Training Procedure of MSADGN}
With the definition of ${\mathcal L}_{\text{inv}}$ and ${\mathcal L}_{\text{spe}}$, the overall loss of MSADGN can be described as follows:
\begin{equation}
\label{equ: L}
    {\mathcal L} = {\mathcal L}_{\text{inv}} + {\mathcal L}_{\text{spe}}
\end{equation}

The optimal parameters of MSADGN can be obtained using the following optimization:
\begin{equation}
\label{equ: optim L}
    ({\hat\theta_f},{\hat\theta_c},{\hat\theta_d}) = \mathop {\arg \min }\limits_{{\theta_{f}},{\theta_{c}},{\theta_{d}}} {\mathcal L}({\theta_{f}},{\theta_{c}},{\theta_{d}})
\end{equation}
where $\theta_f=(\theta_{f_1}, \theta_{f_2})$ represents the parameters of $F_{\text{shared}}$ and $F_{\text{weighted}}$, respectively; $\theta_c=(\theta_{c_1}, \theta_{c_2}, \dots, \theta_{c_K}, \theta_{c_{K+1}})$ denotes the parameters of $C_1$, $C_2$, \dots, $C_K$, and $C_{\text{weighted}}$, respectively; and $\theta_d=(\theta_{d_1}$, $\theta_{d_2}$, $\dots$, $\theta_{d_{K(K-1)/2}})$ denotes the parameters of $D_1$, $D_2$, $\dots$, and $D_{K(K-1)/2}$, respectively.

Finally, according to Eqs.~\eqref{equ: z} and \eqref{equ: w_i}, our model is capable of making real-time predictions for the unseen target domain.
More importantly, the prediction results may benefit from the domain-specific features learned from the multisource domains.
The detailed training procedure of MSADGN is outlined in Algorithm \ref{alg: training procedure of MSADGN}.
We believe that the proposed MSADGN is expected to improve the cross-scene sea\textendash land clutter classification performance.

\begin{algorithm}[!t]\small
\caption{Training procedure of MSADGN.}

\label{alg: training procedure of MSADGN}
\SetAlgoLined
\KwIn {Labeled source data ${\mathcal D}_s^1=\{({\mathbf x}_i^l,y_i^l,d_i^l=1)\}_{i=1}^{n_l}$;

\hspace{28pt}Unlabeled source data ${\mathcal D}_s^{k\neq 1}=\{({\mathbf x}_i^u,d_i^u=k)\}_{i=1}^{n_k}$;

\hspace{28pt}Unlabeled target data ${\mathcal D}_t=\{{\mathbf x}_j^t\}_{j=1}^{n_t}$}

\KwOut{Target category labels ${\mathcal D}_t = \{ y_j^t\} _{j = 1}^{{n_t}}$}

\textbf{// TRAINING}

\textbf{Initialize} Weight parameters: $\theta_f$, $\theta_c$, and $\theta_d$;

\hspace{38pt}Maximum number of training epochs: $N_e$;

\hspace{38pt}Mini-batch size: $m$;

\hspace{38pt}Maximum number of mini-batch: $N_b$;

\For{$n_e \leftarrow 1$ \KwTo $N_e$}
    {\For{$n_b \leftarrow 1$ \KwTo $N_b$}
        {Randomly sample ${\mathcal M}_s^1 = \{({\mathbf x}_i^l,y_i^l,d_i^l=1)\}_{i=1}^{m}$ from \par
        ${\mathcal D}_s^1$;
        
        Randomly sample ${\mathcal M}_s^{k\neq 1} = \{({\mathbf x}_i^u,d_i^u=k)\}_{i=1}^{m}$ from\par ${\mathcal D}_s^{k\neq 1}$;
        
        {\bf // Perform domain-related pseudolabeling.}
        
        Select ${\mathcal M}_{s'}^{k\neq 1} = \{({\mathbf x}_i^u, \widehat{y_i^u}, d_i^u=k)\}_{i=1}^{m'_k}$ by Eq.~\eqref{equ: domain-related pseudolabel},\par
        where $m'_k$ represents the number of pseudolabeled\par
        samples in ${\mathcal M}_{s'}^{k\neq 1}$;

        {\bf // Perform domain-invariant and domain-specific\par
        DGs.}
        
        {\bf Forward:} calculate ${\mathcal L}$ by Eq.~\eqref{equ: L};
        
        {\bf Backward:} update $\theta_f$, $\theta_c$, and $\theta_d$ by:\par        
        \hspace{60pt}$\theta_f\leftarrow \theta_f - \frac{\partial{\mathcal L}}{\partial\theta_f}$
        
        \hspace{60pt}$\theta_c \leftarrow \theta_c - \frac{\partial\mathcal L}{\partial\theta_c}$
        
        \hspace{60pt}$\theta_d \leftarrow \theta_d - \frac{\partial\mathcal L}{\partial\theta_d}$.}
    }
    \textbf{// PREDICTION}\par 
    Classify the target data $\{ {\mathbf x}_j^t\} _{j = 1}^{{n_t}}$ by Eqs.~\eqref{equ: z} and \eqref{equ: w_i}.\par
\textbf{return} $\{ y_j^t\} _{j = 1}^{{n_t}}$
\end{algorithm}

\subsection{Theoretical Insight}
\label{subsec: Theoretical Insight}
In this section, we theoretically demonstrate the rationality of our method using the DG theory proposed in \cite{wang2023generalizing}.

\begin{theorem}
\label{theo: DG}
(Domain generalization error upper bound \cite{wang2023generalizing})
Let $\xi=d_{\mathcal H}({\mathcal D}_t,\bar{\mathcal D}_t)$ denotes the $\mathcal H$-divergence between the unseen target domain ${\mathcal D}_t$ and its nearest neighbor in the convex hull of multisource domains ${\mathcal D}_{s'}$.
Let $\eta$ denotes the largest distribution divergence between ${\mathcal D}_s^i$ and ${\mathcal D}_s^j$, $i\neq j$.
Then, the error on ${\mathcal D}_t$ of hypothesis $h$ is upper-bound by the weighted error on ${\mathcal D}_{s'}$:
\begin{equation}
\label{equ: epsilon}
\begin{aligned}
        \epsilon_t(h) \leq & \sum_{i=1}^{N_{s'}}\pi_i^*\epsilon_{s'}^i(h) +  \frac{\xi+\eta}{2}\\
        & + \min\{{\mathbb E_{\bar{\mathcal D}_t}}[\lvert f_t - f_{s'_\pi}\rvert],{\mathbb E_{{\mathcal D}_t}}[\lvert f_t - f_{s'_\pi}\rvert]\}
\end{aligned}
\end{equation}
where $\min\{{\mathbb E_{\bar{\mathcal D}_t}}[\lvert f_t - f_{s'_\pi}\rvert],{\mathbb E_{{\mathcal D}_t}}[\lvert f_t - f_{s'_\pi}\rvert]\}$ denotes the different labeling functions between $f_t$ and $f_{s'_\pi}$.
\end{theorem}

In the context of the cross-scene sea\textendash land clutter classification problem, we consider that the label space of ${\mathcal D}_s$ (or ${\mathcal D}_{s'}$) and ${\mathcal D}_t$ is the same, and the distribution difference between the two is not particularly large.
This aligns approximately with the covariate shift assumption, suggesting that variations in the input multisource domain distribution do not significantly impact the target domain label distribution.
Thus, it can be inferred that $\xi$ and the labeling function error ($\min\{\cdot,\cdot\}$) are relatively small.

On the other hand, the risk $\eta$ is bounded by Eq.~\eqref{equ: L_{inv}} of the domain-invariant module. In addition, the weighted error $\sum_{i=1}^{N_{s'}}\pi_i^*\epsilon_{s'}^i(h)$ on ${\mathcal D}_{s'}$ is bounded by Eqs.~\eqref{equ: z} and \eqref{equ: w_i} of the domain-specific module.
Therefore, our MSADGN can be guaranteed by Theorem~\ref{theo: DG}.

\section{Experiments and Analysis}
\label{sec: Experiments and Analysis}
In this section, extensive experiments are conducted to validate the effectiveness and superiority of our proposed MSADGN.
First, we introduce two benchmark datasets: CS-SLCS and CS-HRRSI, along with experimental implementation details.
The experiments on CS-HRRSI aim to demonstrate the generality of MSADGN.
Second, we compare MSADGN with state-of-the-art methods to verify its superiority.
Third, an ablation study is conducted to demonstrate the effectiveness of MSADGN.
Finally, further empirical analysis is performed for our method.

\subsection{Experimental Settings}
\label{subsec: Experimental Settings}
\subsubsection{Dataset Preparation}
\label{subsubsec: Dataset Preparation}
\begin{figure}[!t]
\centering
\subfloat[\bf R]{
\fcolorbox{mycolor}{white}{
\includegraphics[width=1in]{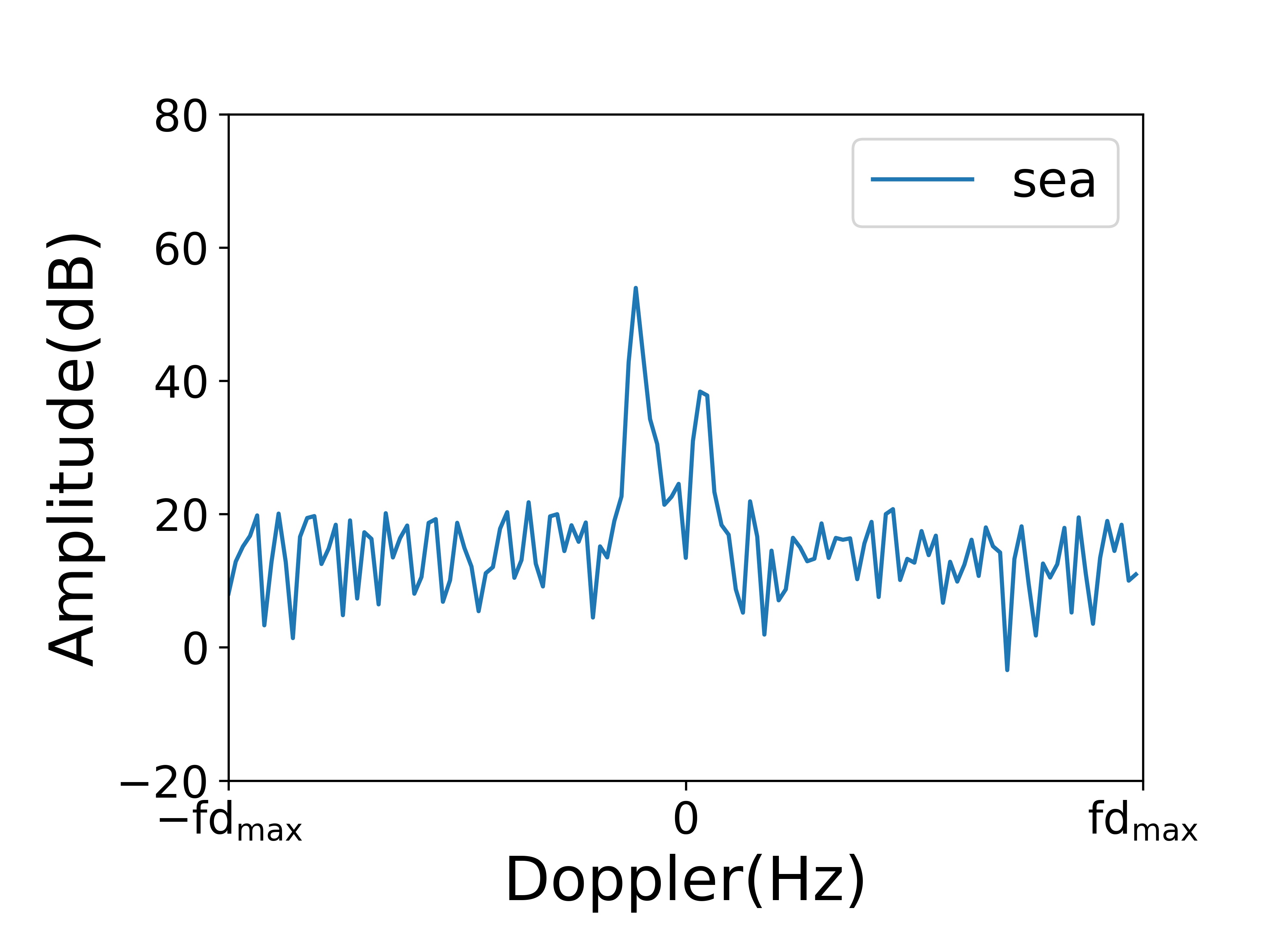}
\includegraphics[width=1in]{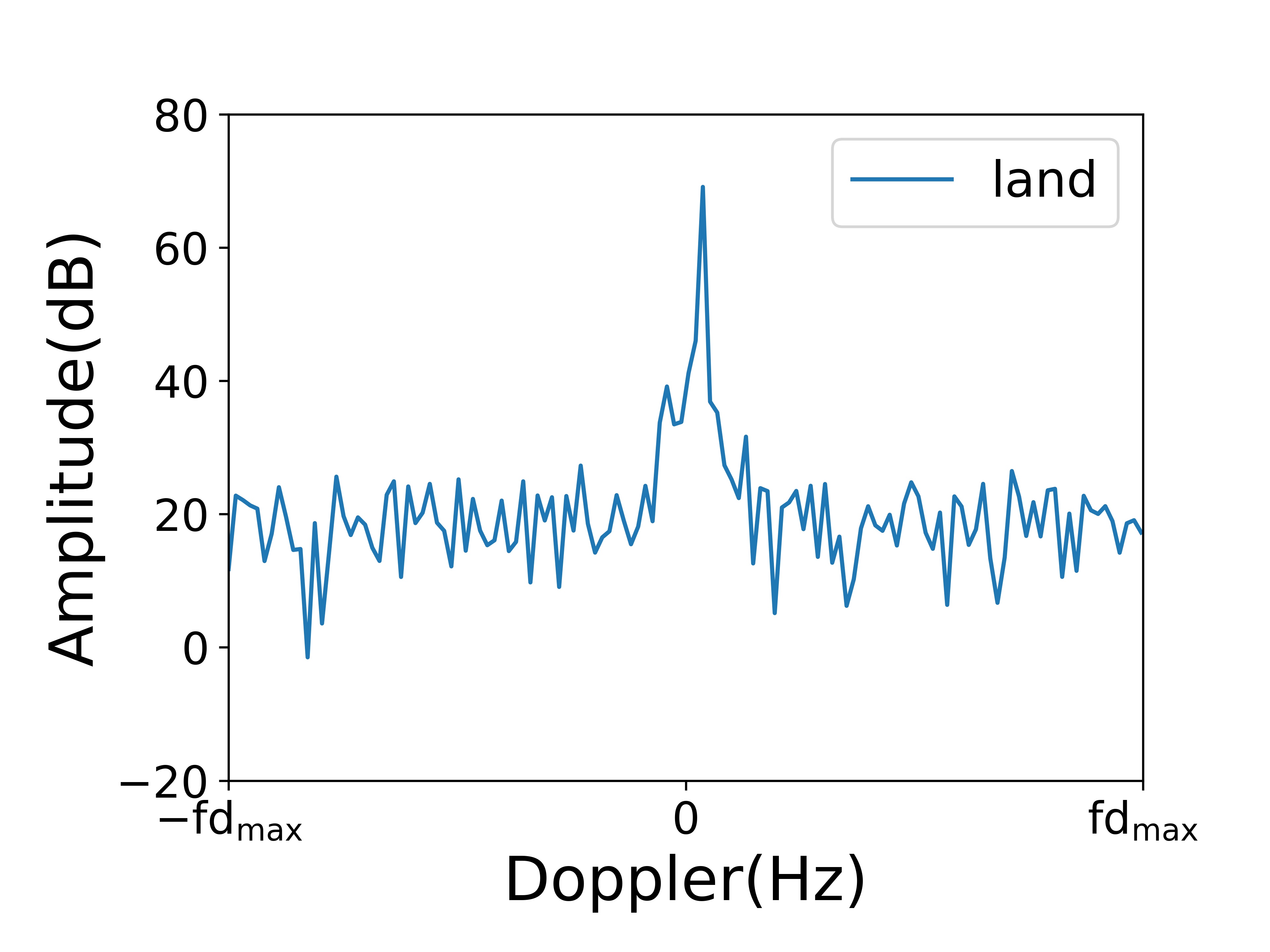}
\includegraphics[width=1in]{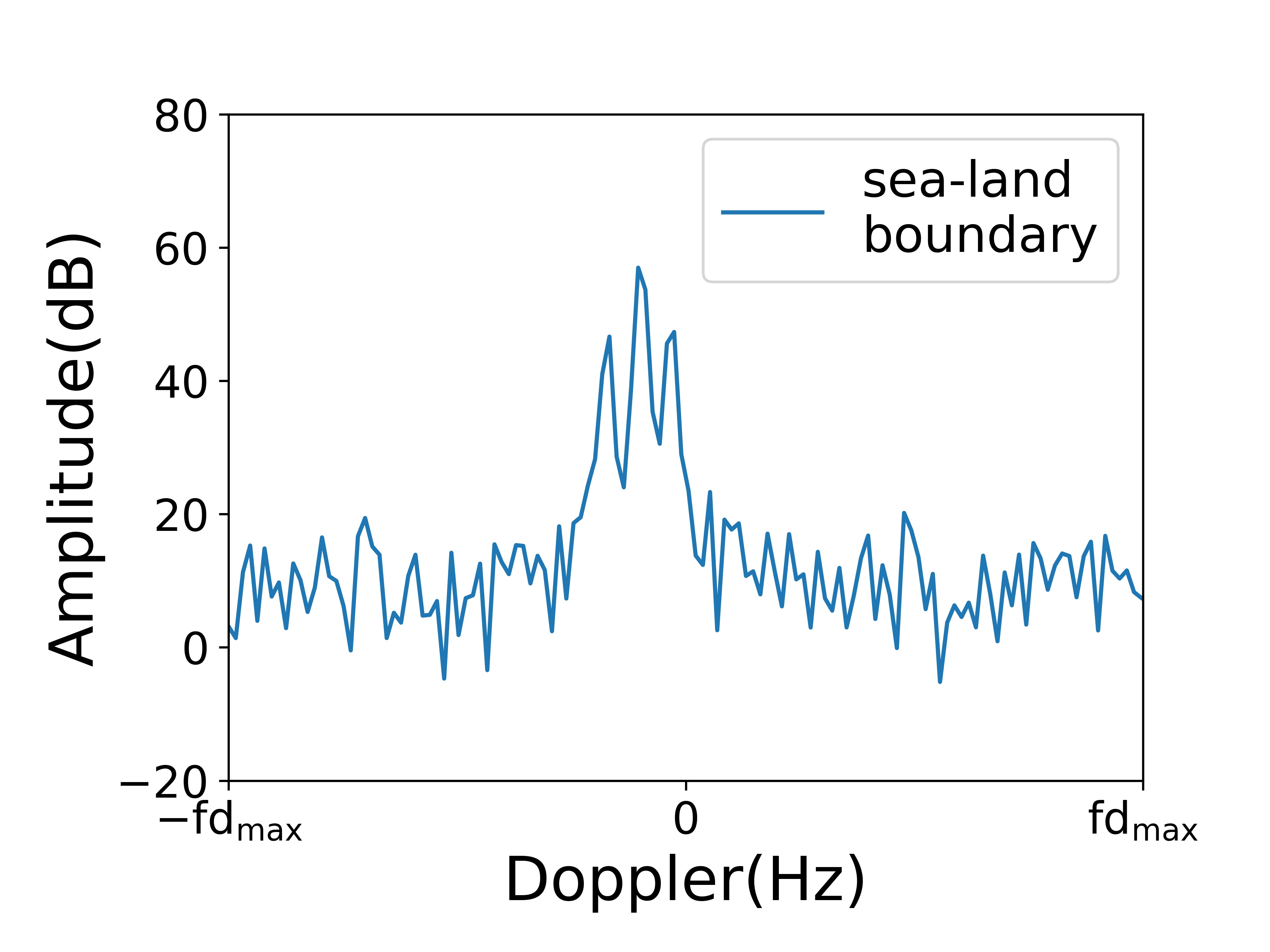}
}}
\vspace{0pt}
\subfloat[\bf S]{
\fcolorbox{mycolor}{white}{
\includegraphics[width=1in]{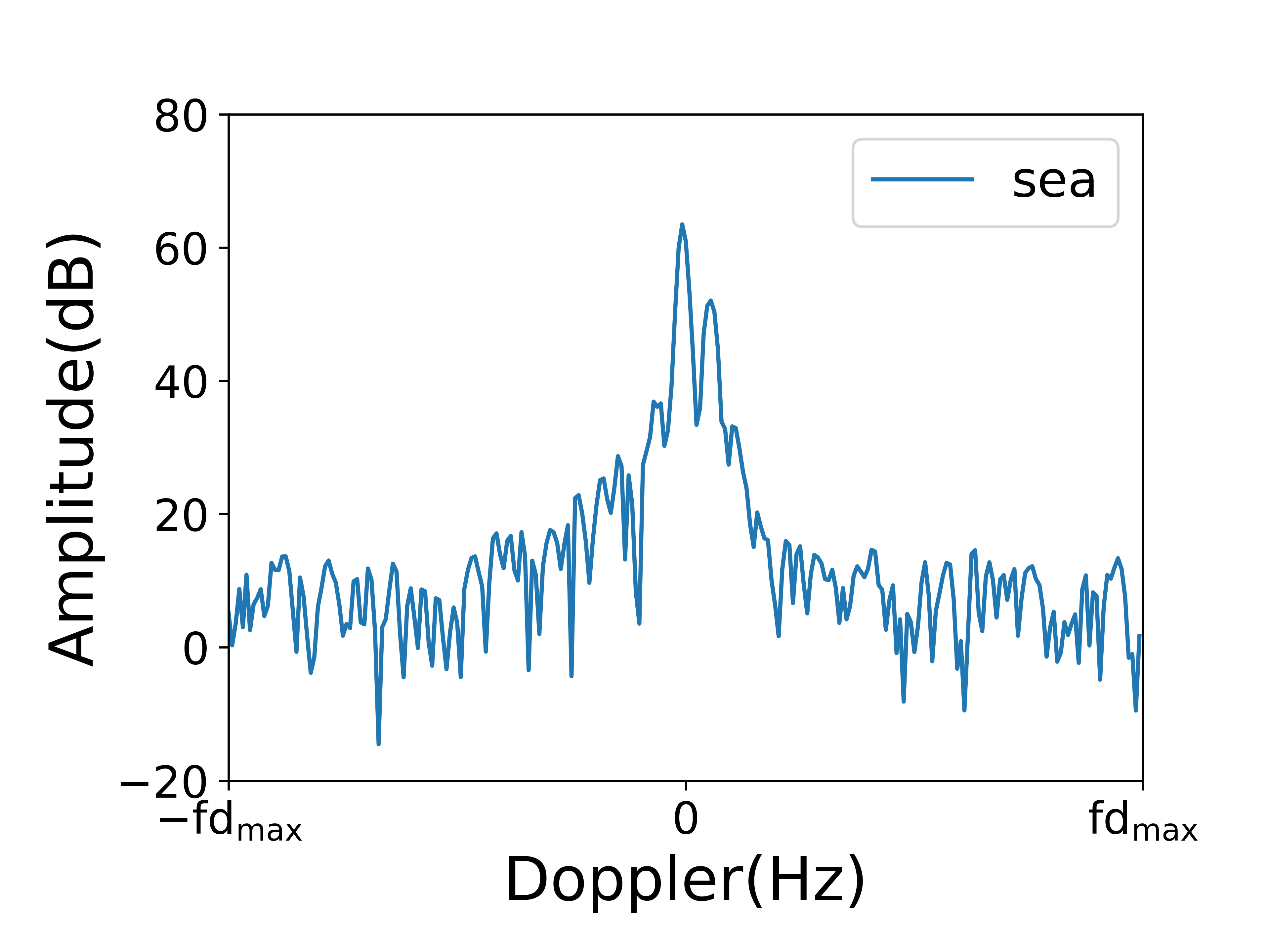}
\includegraphics[width=1in]{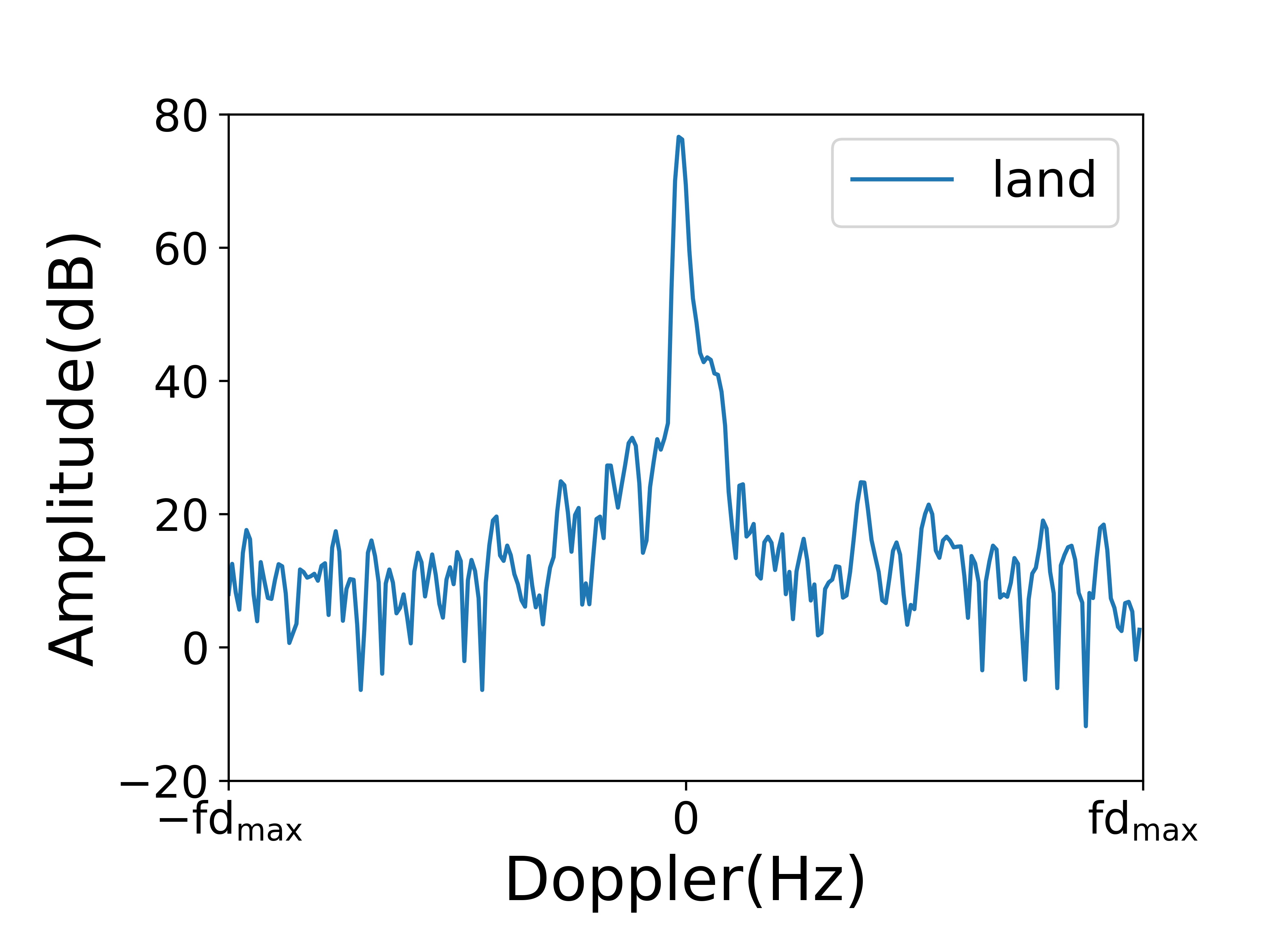}
\includegraphics[width=1in]{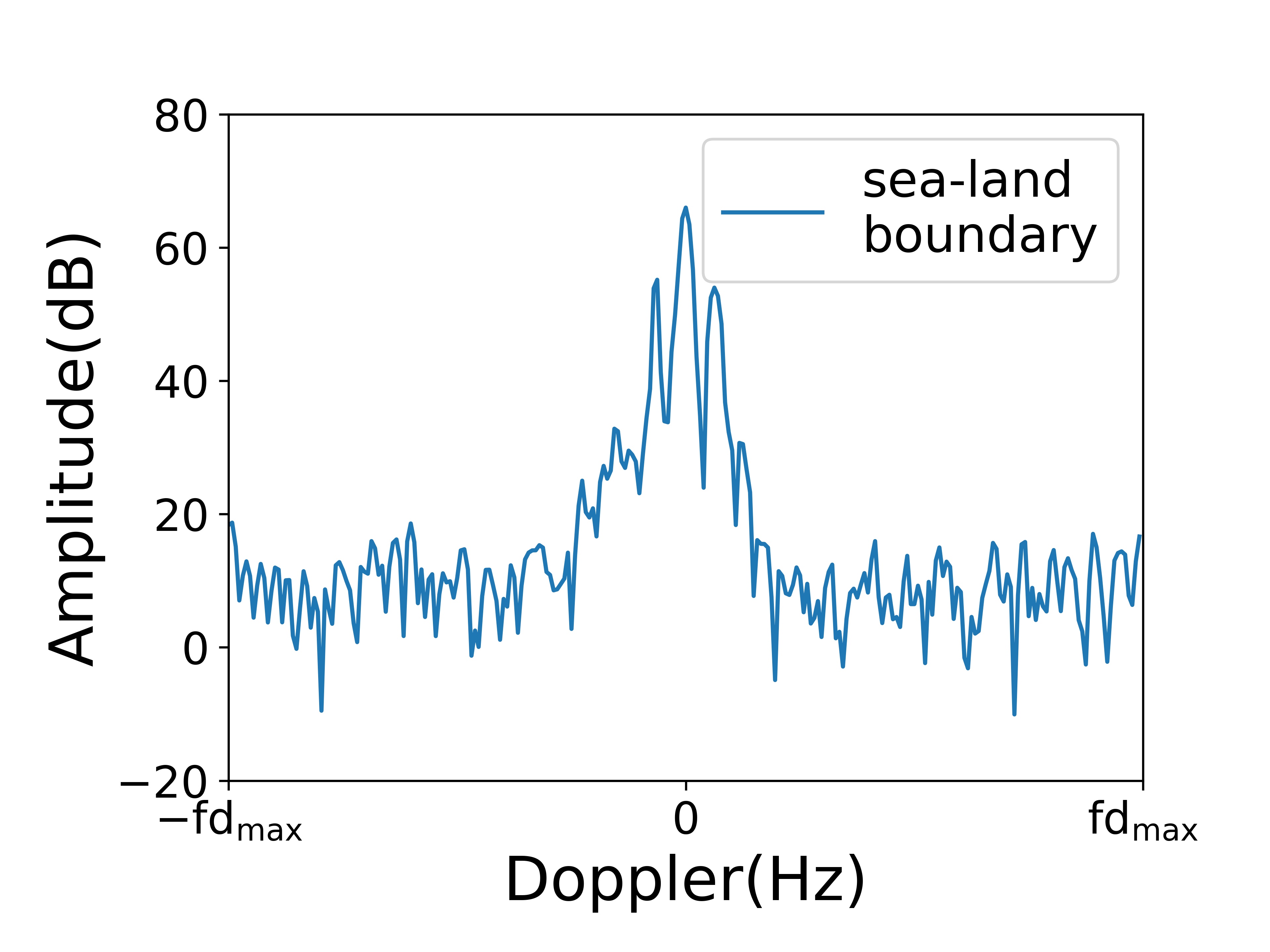}
}}
\vspace{0pt}
\subfloat[\bf M]{
\fcolorbox{mycolor}{white}{
\includegraphics[width=1in]{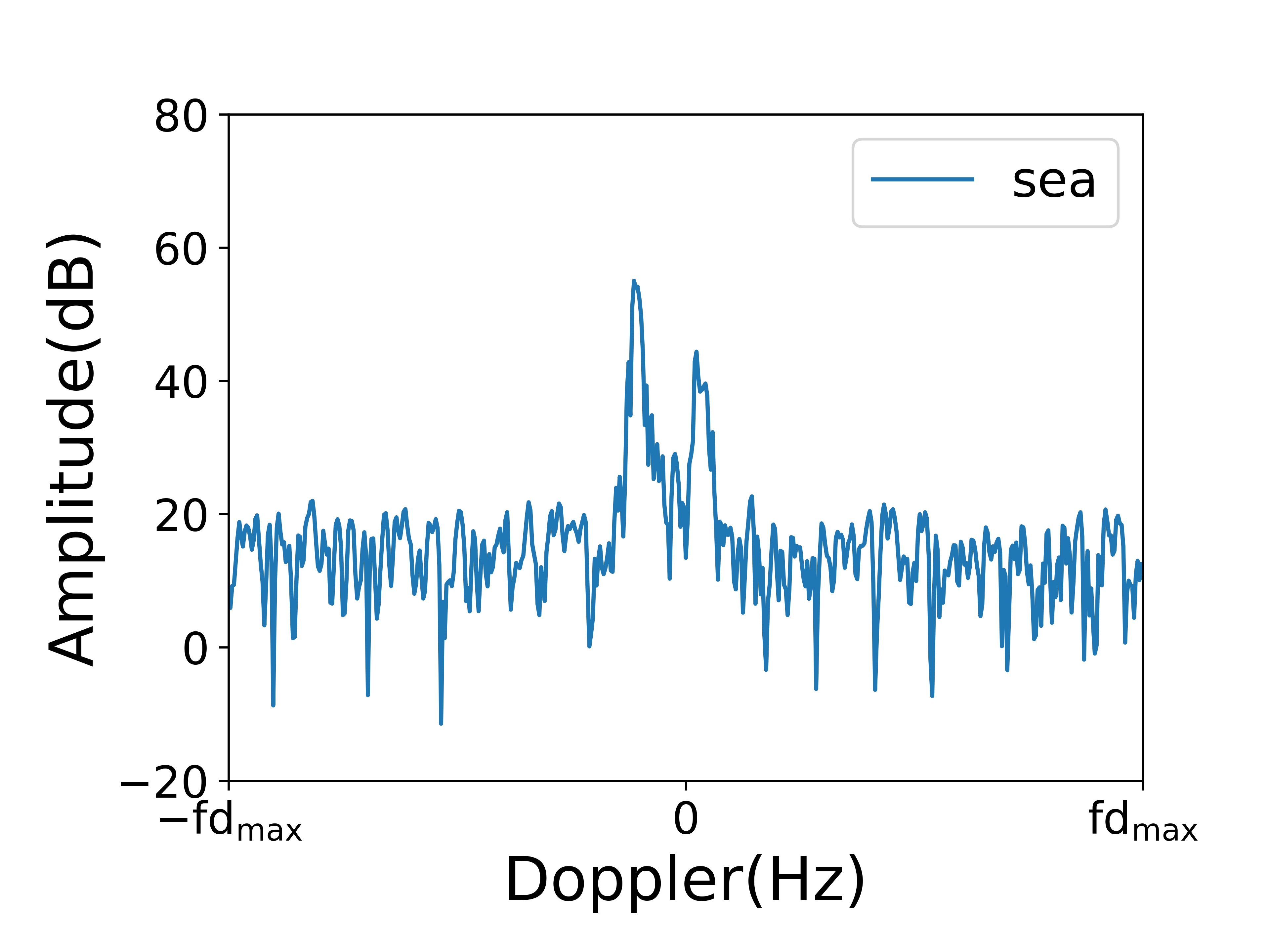}
\includegraphics[width=1in]{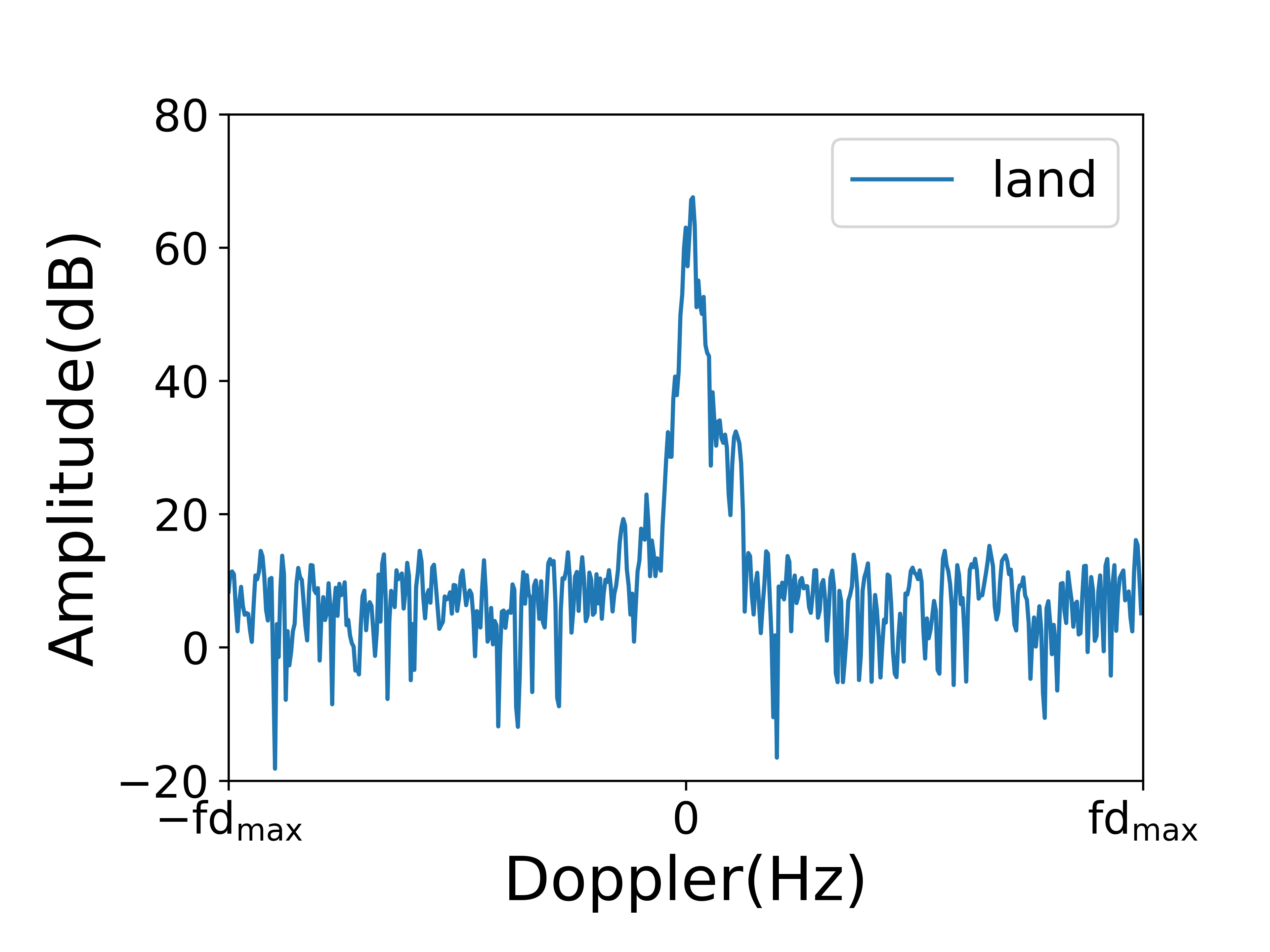}
\includegraphics[width=1in]{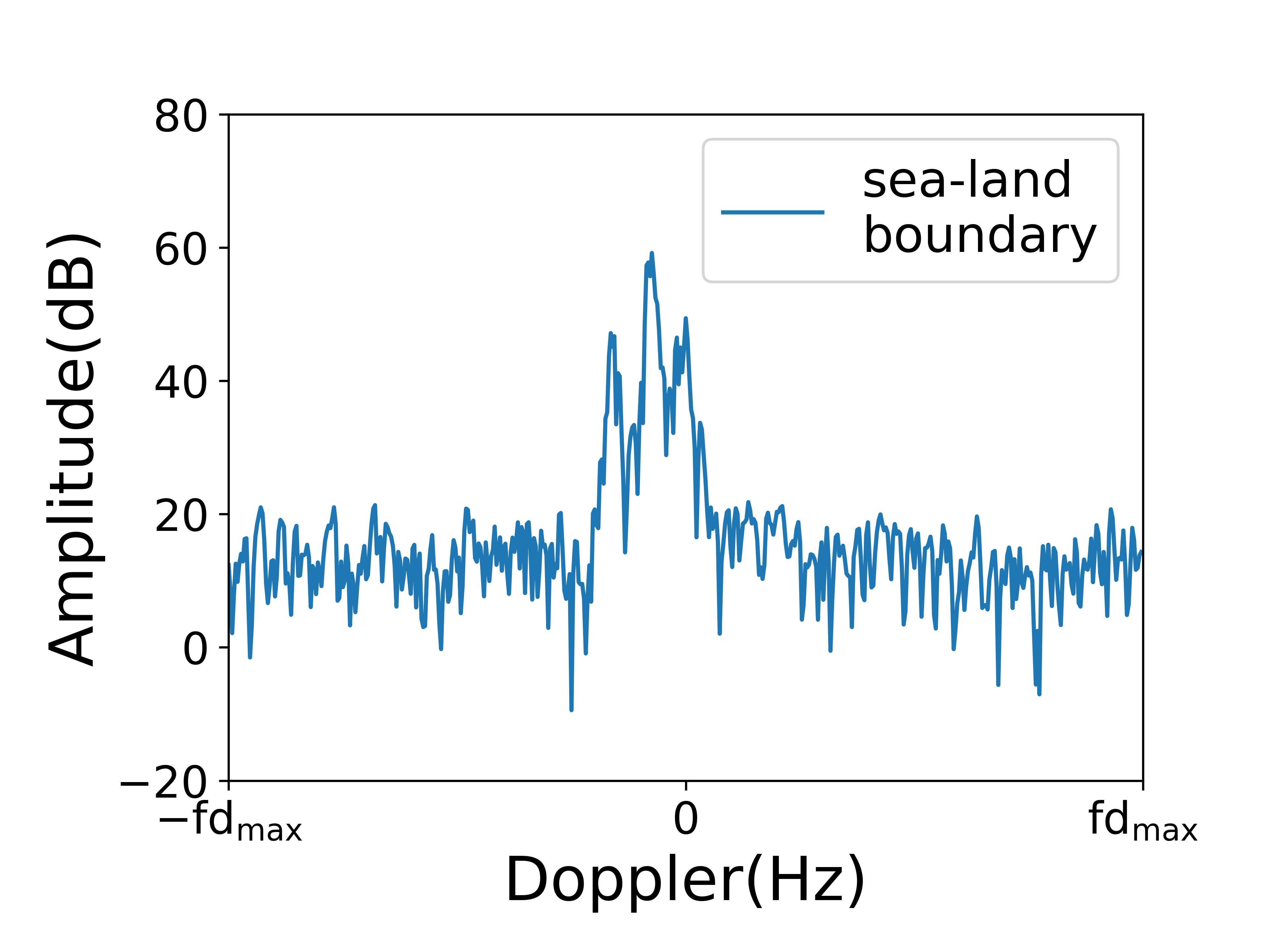}
}}
\vspace{0pt}
\subfloat[\bf L]{
\fcolorbox{mycolor}{white}{
\includegraphics[width=1in]{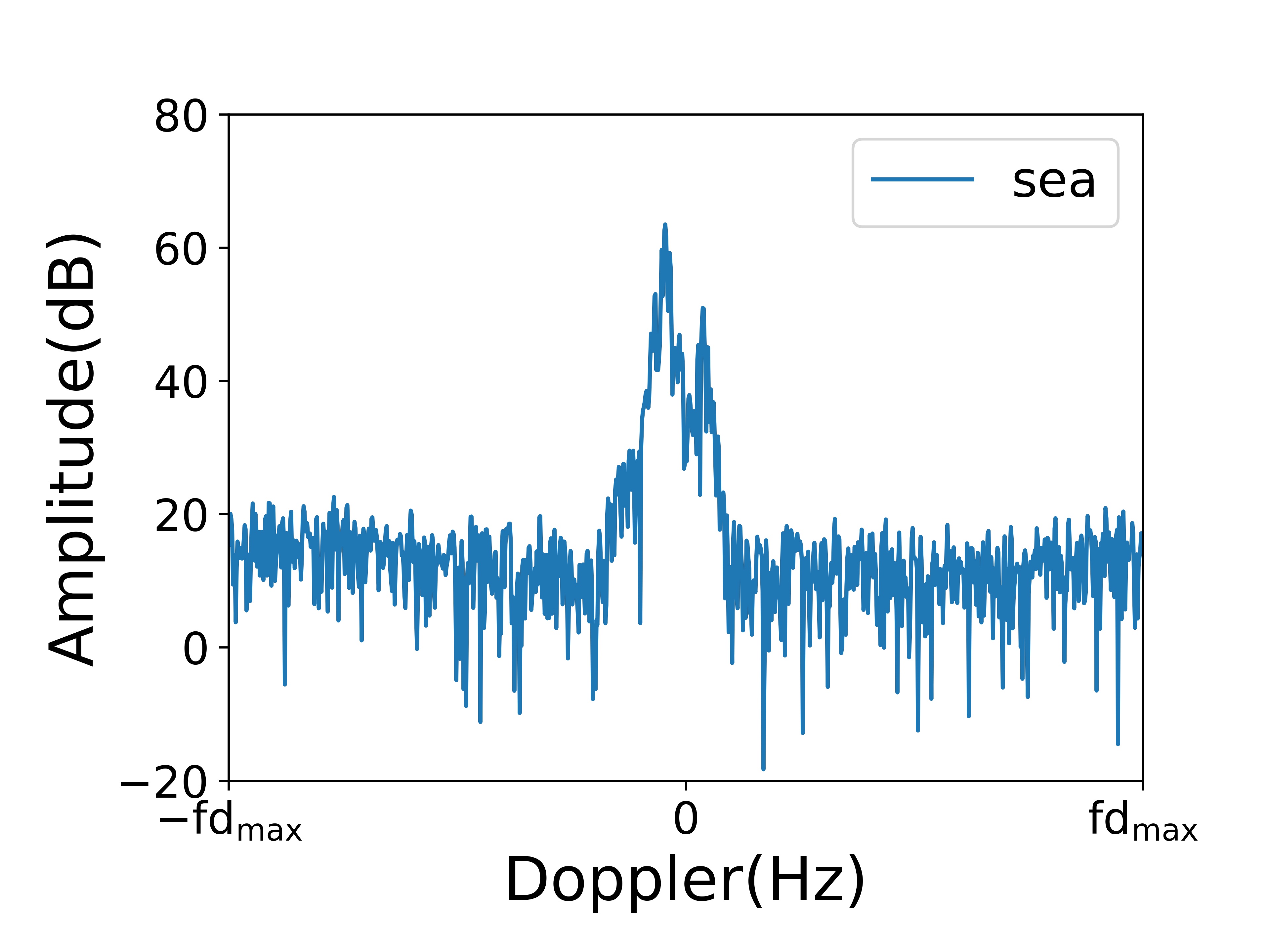}
\includegraphics[width=1in]{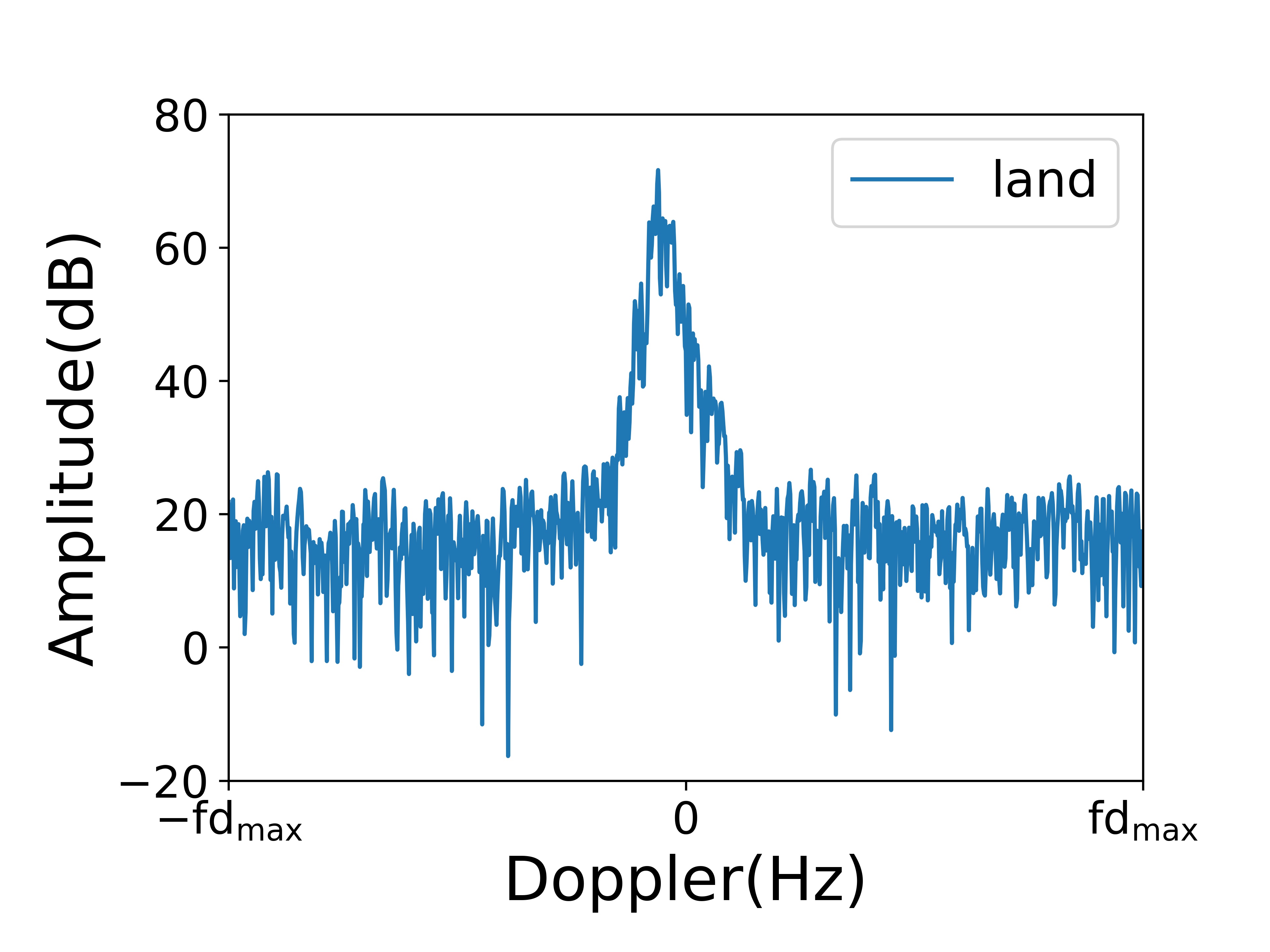}
\includegraphics[width=1in]{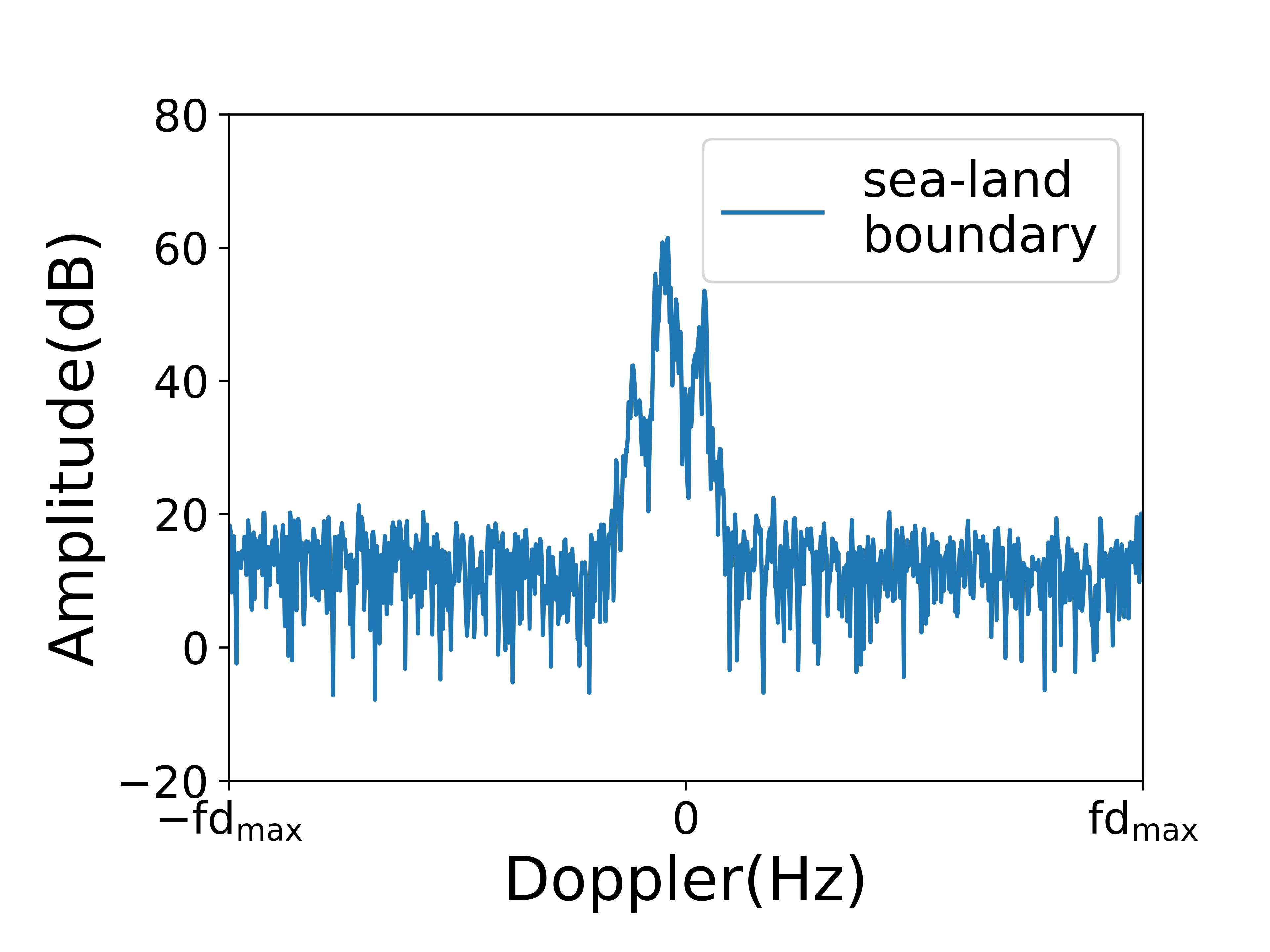}
}}
\caption{Samples of CS-SLCS. Order of samples from left to right: 1) sea clutter; 2) land clutter; and 3) sea\textendash land boundary clutter.}
\label{fig: CS-SLCS}
\end{figure}

See Fig.~\ref{fig: CS-SLCS}, CS-SLCS is the clutter spectrum obtained from OTHR, which contains three categories of signals (1000 sea clutter, 1000 land clutter, and 1000 sea\textendash land boundary clutter) from four domains defined by different coherent integration numbers (CINs): 128 CIN, 256 CIN, 512 CIN, and 1024 CIN.
Please refer to our previous work \cite{zhang2023triple} for a detailed description of CS-SLCS, where CS-SLCS with 256 CIN, 512 CIN, and 1024 CIN is utilized.
Here, to comprehensively validate the performance of MSADGN, we add the domain 128 CIN.
For convenience, 128 CIN, 256 CIN, 512 CIN, and 1024 CIN are abbreviated as {\bf R}, {\bf S}, {\bf M}, and {\bf L}, respectively.
Besides, the dimension of signals from all domains is resized to $1\times 512$ to match the model's input requirements.

\begin{figure}[!t]
\centering
\subfloat[{\bf N}]{
\fcolorbox{mycolor}{white}{
\begin{minipage}[t]{0.1\textwidth}
\centering
\includegraphics[width=0.5in]{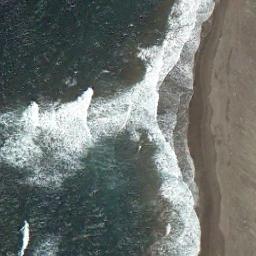}\\
\vspace{1pt}
\includegraphics[width=0.5in]{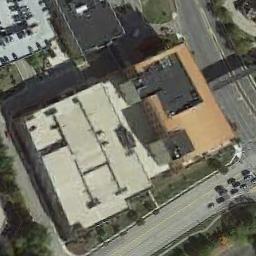}\\
\vspace{1pt}
\includegraphics[width=0.5in]{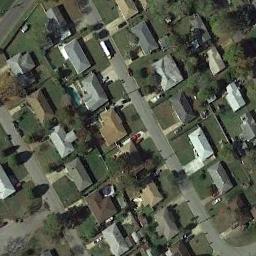}\\
\vspace{1pt}
\includegraphics[width=0.5in]{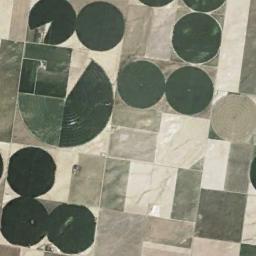}\\
\vspace{1pt}
\includegraphics[width=0.5in]{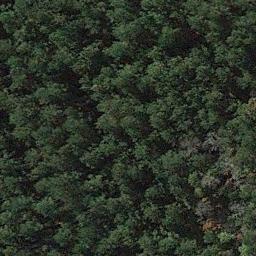}\\
\vspace{1pt}
\includegraphics[width=0.5in]{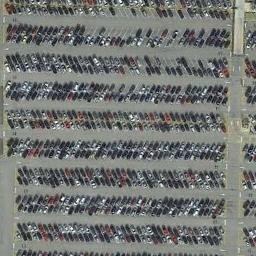}\\
\vspace{1pt}
\includegraphics[width=0.5in]{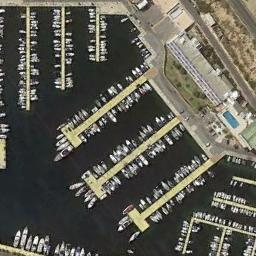}\\
\vspace{1pt}
\includegraphics[width=0.5in]{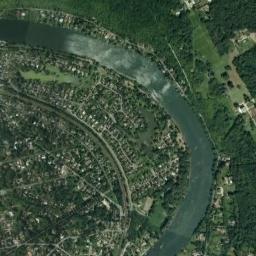}\\
\vspace{1pt}
\includegraphics[width=0.5in]{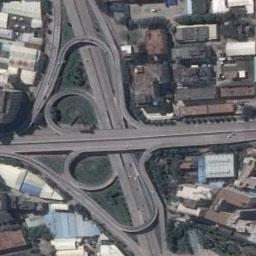}\\
\label{fig: img_s1}
\end{minipage}}}
\subfloat[{\bf A}]{
\fcolorbox{mycolor}{white}{\begin{minipage}[t]{0.1\textwidth}
\centering
\includegraphics[width=0.5in]{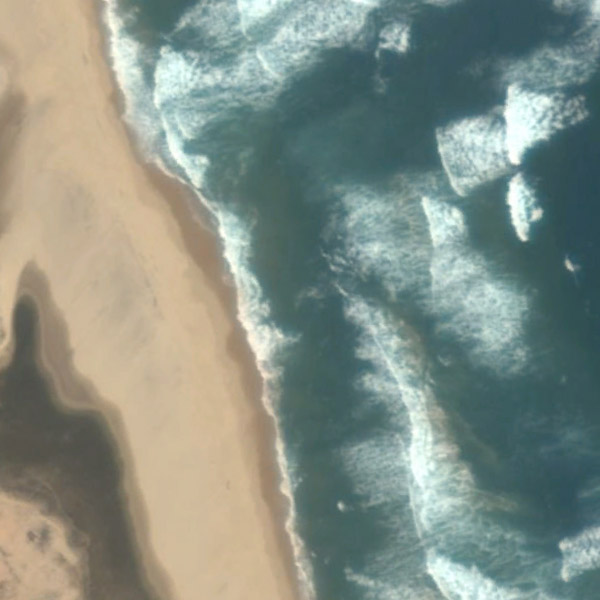}\\
\vspace{1pt}
\includegraphics[width=0.5in]{figure/image/commercial_A.jpg}\\
\vspace{1pt}
\includegraphics[width=0.5in]{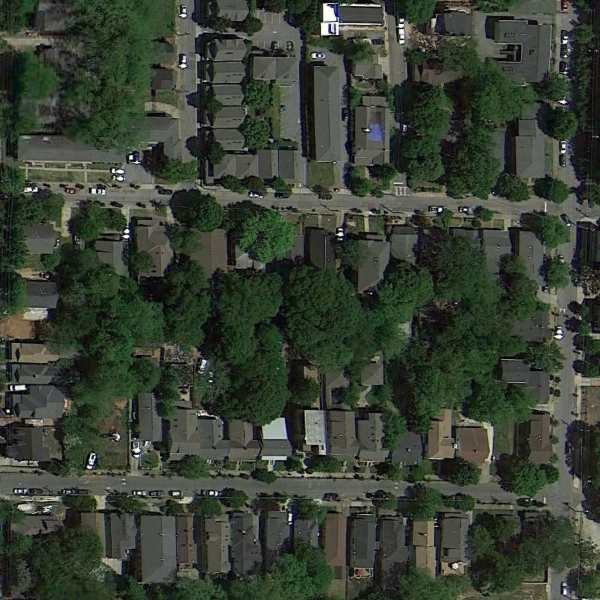}\\
\vspace{1pt}
\includegraphics[width=0.5in]{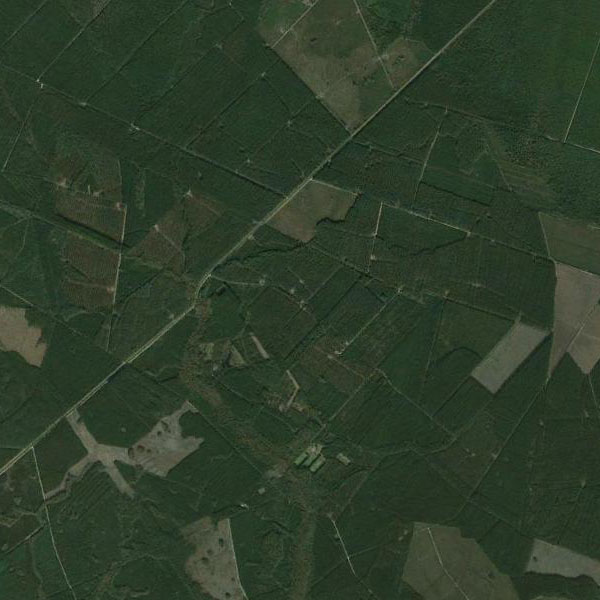}\\
\vspace{1pt}
\includegraphics[width=0.5in]{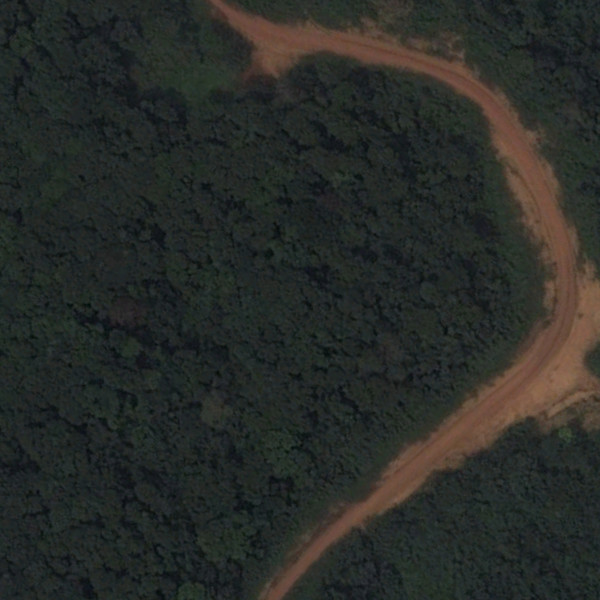}\\
\vspace{1pt}
\includegraphics[width=0.5in]{figure/image/parking_A.jpg}\\
\vspace{1pt}
\includegraphics[width=0.5in]{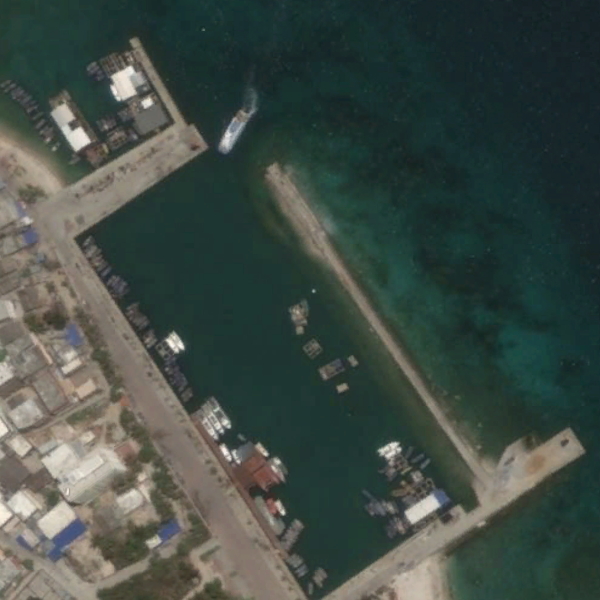}\\
\vspace{1pt}
\includegraphics[width=0.5in]{figure/image/river_A.jpg}\\
\vspace{1pt}
\includegraphics[width=0.5in]{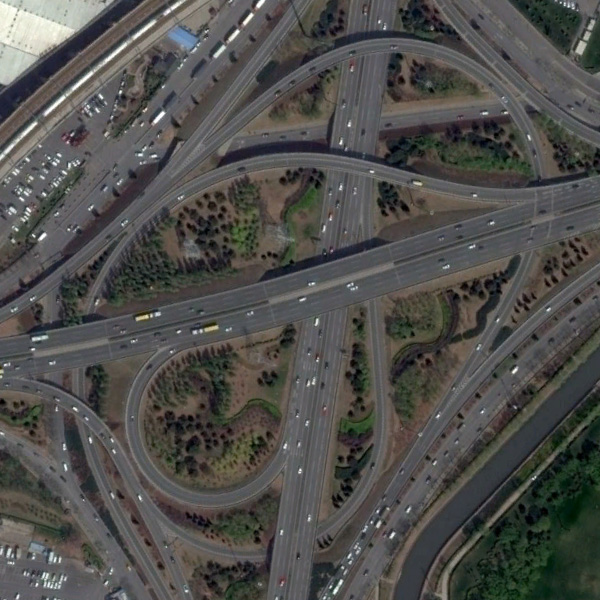}\\
\label{fig: img_s2}
\end{minipage}}}
\subfloat[{\bf U}]{
\fcolorbox{mycolor}{white}{\begin{minipage}[t]{0.1\textwidth}
\centering
\includegraphics[width=0.5in]{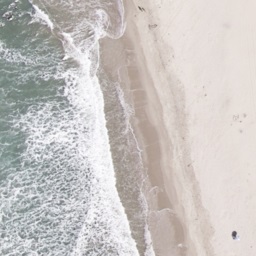}\\
\vspace{1pt}
\includegraphics[width=0.5in]{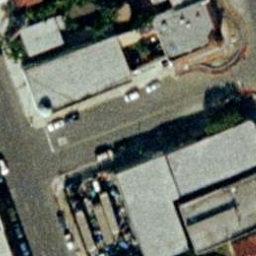}\\
\vspace{1pt}
\includegraphics[width=0.5in]{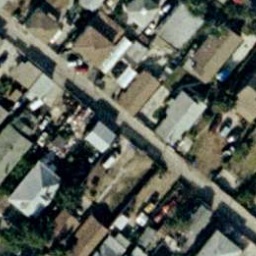}\\
\vspace{1pt}
\includegraphics[width=0.5in]{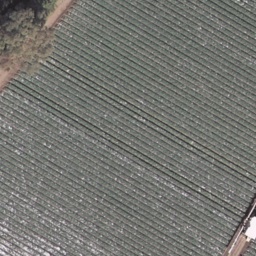}\\
\vspace{1pt}
\includegraphics[width=0.5in]{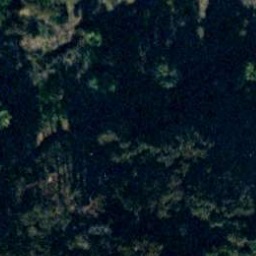}\\
\vspace{1pt}
\includegraphics[width=0.5in]{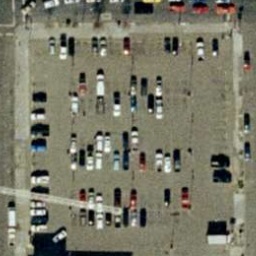}\\
\vspace{1pt}
\includegraphics[width=0.5in]{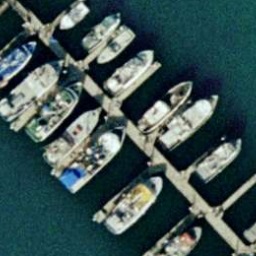}\\
\vspace{1pt}
\includegraphics[width=0.5in]{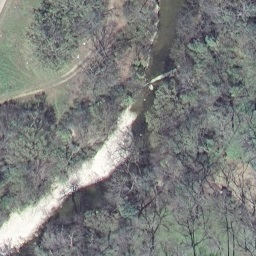}\\
\vspace{1pt}
\includegraphics[width=0.5in]{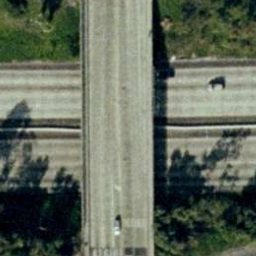}\\
\label{fig: img_s3}
\end{minipage}}}
\subfloat[{\bf W}]{
\fcolorbox{mycolor}{white}{\begin{minipage}[t]{0.1\textwidth}
\centering
\includegraphics[width=0.5in]{figure/image/beach_W.jpg}\\
\vspace{1pt}
\includegraphics[width=0.5in]{figure/image/commercial_W.jpg}\\
\vspace{1pt}
\includegraphics[width=0.5in]{figure/image/residential_W.jpg}\\
\vspace{1pt}
\includegraphics[width=0.5in]{figure/image/farmland_W.jpg}\\
\vspace{1pt}
\includegraphics[width=0.5in]{figure/image/forest_W.jpg}\\
\vspace{1pt}
\includegraphics[width=0.5in]{figure/image/parking_W.jpg}\\
\vspace{1pt}
\includegraphics[width=0.5in]{figure/image/port_W.jpg}\\
\vspace{1pt}
\includegraphics[width=0.5in]{figure/image/river_W.jpg}\\
\vspace{1pt}
\includegraphics[width=0.5in]{figure/image/overpass_W.jpg}\\
\label{fig: img_s4}
\end{minipage}}}

\caption{Samples of CS-HRRSI. Order of samples from top to bottom: 1) beach; 2) commercial; 3) residential; 4) farmland; 5) forest; 6) parking; 7) port; 8) river; and 9) overpass.}
\label{fig: CS-HRRSI}
\end{figure}

See Fig.~\ref{fig: CS-HRRSI}, CS-HRRSI contains nine categories of images (beach, commercial, residential, farmland, forest, parking, port, river, and overpass) from four domains: NWPU-RESISC45 \cite{cheng2017remote}, AID \cite{xia2017aid}, UC Merced \cite{yang2010bag}, and WHU-RS19 \cite{xia2010Structural}.
Please refer to our previous work \cite{zhang2023triple} for a detailed description of CS-HRRSI, where CS-HRRSI with NWPU-RESISC45, AID, and UC Merced is utilized.
Here, to comprehensively validate the performance of MSADGN, we add the domain WHU-RS19.
For convenience, NWPU-RESISC45, AID, UC Merced and WHU-RS19 are abbreviated as {\bf N}, {\bf A}, {\bf U} and {\bf W}, respectively.
Besides, the dimension of images from all domains is resized to $224\times 224$ pixels to match the model's input requirements.

\subsubsection{Implementation Details}
\label{subsubsec: Implementation Details}
In our experiment, we adopt the DCNN architecture used in \cite{zhang2023triple} as the backbone of MSADGN and all comparison methods.
In other words, in the CS-SLCS task, the feature extractor is composed of four 1-D convolutional (Conv1d) layers, the classifier or discriminator is composed of three FC layers, and Gaussian weight initialization is applied to the feature extractor, classifier, and discriminator to stabilize training; in the CS-HRRSI task, the feature extractor is composed of ResNet-50 pretrained on the ImageNet dataset, the classifier or discriminator consists of three FC layers, and Gaussian weight initialization is applied to the classifier and discriminator to stabilize training.
The only difference is that the output dimension of $C_{\text{weighted}}$ for these two tasks is changed from $C$ to $K$.
Besides, we set the batch size to 32 for each domain and employ the Adam optimizer \cite{kingma2014adam} with an initial learning rate of 0.0001 and a decay rate of 0.5 every 10 epochs.
The tradeoff parameter $\alpha$ in Eq.~\eqref{equ: Phi} is set to 0.2.
To ensure reliable results, we conduct five experiments and report the mean classification accuracy after 50 epochs.
All experiments are conducted using the PyTorch DL library \cite{paszke2019pytorch} on an NVIDIA GeForce RTX 3090 GPU.

\subsection{Comparisons with State-of-the-Art methods}
\label{sub: Comparisons with State-of-the-Art methods}
Abundant state-of-the-art methods are selected to verify the superiority of MSADGN.
These methods are categorized into three groups: one baseline (BL) method, five DA-based DG methods, and five pure SSDG methods.
For fairness in comparative results, all methods adopt consistent datasets, hyperparameters, and submodules, as detailed in Section~\ref{subsubsec: Implementation Details}.
The comparison methods are listed as follows.
\begin{itemize}
    \item[1)] \textit{BL}.
    DCNN without DG \cite{he2016deep} (also called ERM) is used as the BL method, meaning only the labeled source domain is used to train DCNN.

    \item[2)] \textit{DA}.
    The selected DA-based DG methods include deep domain confusion (DDC) \cite{tzeng2014deep}, deep adaptation network (DAN) \cite{long2015learning}, DANN \cite{ganin2015unsupervised}, error-correcting boundaries mechanism with feature adaptation metric (ECB-FAM) \cite{ma2021unsupervised}, and TLADAN \cite{zhang2023triple}.
    Among these, DDC and DAN can be considered as discrepancy-based methods; DANN can be regarded as an adversarial-based method; and ECB-FAM and TLADAN can be viewed as discrepancy- and adversarial-based methods.
    In this group, the model is trained using both the labeled source domain and the unlabeled source domain (by combining all unlabeled source domains into one domain).

    \item[3)] \textit{DSDGN}.
    It is an SSDG method in fault diagnosis.
    Due to the commonality of 1-D signals in both sea\textendash land clutter and fault diagnosis data, it is also applicable to CS-SLCS.
    DSDGN integrates the traditional pseudolabel method into DANN, enabling the realization of SSDG with an accessible labeled source domain and an accessible unlabeled source domain.
    In the context of multisource SSDG, the discriminator of DSDGN is expanded to $K(K-1)/2$ to distinguish the features of any two source domains.
    Since our MSADGN is an improved version of DSDGN, it can be viewed as a BL method at the DG level.
    
    \item[4)] \textit{DFGN}.
    It is also an SSDG method in fault diagnosis and employs the pseudolabel technique.
    Different from DSDGN, domain fuzzy generalization network (DFGN) \cite{ren2023domain} utilizes pseudolabel to compute the center loss and expands the output dimension of the discriminator into $K$ to achieve multisource SSDG.

    \item[5)] \textit{BPL}.
    It is an improved version of DSDGN in CV.
    On the one hand, the traditional pseudolabel is improved to a domainaware pseudolabeling module in better pseudolabel (BPL) \cite{wang2023better}.
    Note that the difference between this module and our domain-related pseudolabeling module lies in the definition of $\psi({\mathbf x}_i^u)$.
    The former computes ${\mathbf M}_1$ using ${\mathcal D}_s^{k\neq 1}$, while the latter computes ${\mathbf M}_1$ using ${\mathcal D}_s^1$.
    On the other hand, the mixup data augmentation technique is applied to generate additional domains and a dual classifier is integrated into the network to enhance the model's generalization capability.
    
    \item[6)] \textit{ADGN}.
    It is also an SSDG method in fault diagnosis.
    Note that the adversarial DG network (ADGN) \cite{li2022new} can be viewed as an improved version of the DA via a task-specific classifier (DATSNET) \cite{zheng2022domain}, which is a DA method in remote sensing.
    Specifically, the total discrepancy loss of the dual classifier is the sum of discrepancy losses computed from each unlabeled source domain.
    
    \item[7)] \textit{ECB-FAM-DG}.
    To the best of our knowledge, there is no SSDG method in remote sensing, particularly in OTHR.
    Similar to the way ADGN improves DATSNET, we improve ECB-FAM for DG (ECB-FAM-DG).
    On the one hand, the total discrepancy loss of the three classifiers is also the sum of discrepancy losses computed from each unlabeled source domain.
    On the other hand, in the FAM structure, the shallow common features are extracted from multisource domains.
    
\end{itemize}

Table.~\ref{table: Classification Accuracy on CS-SLCS} lists the experimental results on CS-SLCS.
From the average results, it can be seen that all DA-based DG methods are better than the BL method (by about 5\%-8\%) and worse than the pure DG methods (by about 2\%-7\%).
The reason for the former is that they attempt to align features between the labeled source domain and the unlabeled source domain, enabling the learned features to generalize well to the unseen target domain.
The reason for the latter is that, although feature alignment is performed, the unlabeled source domains are considered together, thus failing to adequately extract domain-invariant features between multisource domains.
Further, MSADGN performs the best (about 12\% higher than BL).
We suppose the reason is that MSADGN can extract excellent domain-specific features.
Besides, MSADGN achieves the best or second-best results in each DG task.
This validates the superiority of our method.

Table.~\ref{table: Classification Accuracy on CS-HRRSI} lists the experimental results on CS-HRRSI.
From the average results, it can be seen that all DA-based DG methods are better than the BL method (by about 1\%-4\%) and worse than the pure DG methods (by about 1\%-5\%).
Besides, MSADGN performs the best (about 6\% higher than BL).
Due to the increased complexity of DG tasks on CS-HRRSI compared to CS-SLCS, the accuracy improvement of all DA-based DG methods and pure DG methods on CS-HRRSI is not as obvious as the improvement observed on CS-SLCS.
However, a consistent conclusion can be drawn that pure DG methods outperform DA-based DG methods, with MSADGN exhibiting optimal performance.
Besides, MSADGN achieves the best or second-best results in most DG tasks.
This further validates the superiority of our method.

\begin{table*}[!t]
\centering
\caption{Classification Accuracy (\%) with State-of-the-Art Methods on CS-SLCS (The Title in The First Row Denotes ${\mathcal D}_t$ and in The Second Row Denotes ${\mathcal D}_s^1$, and The Best Results Are in Bold and Second-Best Results Are Underlined)}
\label{table: Classification Accuracy on CS-SLCS}
\centering
\resizebox{\linewidth}{!}{
\begin{tabular}{llccccccccccccc}
\toprule
   \multicolumn{2}{c}{\bf Method} & \multicolumn{3}{c}{\bf R} & \multicolumn{3}{c}{\bf S} & \multicolumn{3}{c}{\bf M} & \multicolumn{3}{c}{\bf L} & {\bf Avg} \\
   \cmidrule(r){3-5} \cmidrule(r){6-8}
   \cmidrule(r){9-11}
   \cmidrule(r){12-14}
    & & {\bf S} & {\bf M} & {\bf L} & {\bf R} & {\bf M} & {\bf L} & {\bf R} & {\bf S} & {\bf L} & {\bf R} & {\bf S} & {\bf M} \\
\midrule
{\bf BL} & {\bf ERM~\cite{he2016deep}} & 89.01 & 82.37 & 73.40 & 81.33 & 83.83 & 73.90 & 76.71 & 77.39 & 87.48 & 68.11 & 78.23 & 92.56 & 80.36\\
\hline
\multirow{6}*{\bf DA} & {\bf DDC~\cite{tzeng2014deep}} & 92.43 & 85.58 & 80.00 & 87.39 & 89.50 & 81.32 & 82.11 & 85.46 & 89.02 & 80.25 & 82.72 & 94.44 & 85.85\\
& {\bf DAN~\cite{long2015learning}} & 92.67 & 86.31 & 81.62 & 87.21 & 88.58 & 81.57 & 83.98 & 86.82 & 88.53 & 81.40 & 81.18 & 93.97 & 86.15\\
& {\bf DANN~\cite{ganin2015unsupervised}} & 92.99 & 88.05 & 83.50 & 88.98 & 90.34 & 82.73 & 84.03 & 87.29 & 89.31 & 83.17 & 82.80 & 94.22 & 87.28\\
&{\bf ECB-FAM~\cite{ma2021unsupervised}} & 92.91 & 87.44 & 84.02 & 89.70 & 90.59 & 83.10 & 85.20 & 88.18 & 89.02 & 83.32 & 83.11 & 93.59 & 87.52\\
& {\bf TLADAN~\cite{zhang2023triple}} & 93.32 & 88.06 & 83.88 & 89.99 & 90.82 & 84.19 & 86.30 & 89.52 & 89.35 & 84.01 & 83.38 & 94.27 & 88.09\\
\hline
\multirow{6}*{\bf DG} & {\bf DSDGN~\cite{liao2020deep}} & 94.13 & \bf{91.52} & 85.87 & 92.01 & 93.30 & 86.00 & \underline{88.56} & 92.21 & 90.77 & 86.87 & 85.49 & 95.93 & 90.22\\
& {\bf DFGN~\cite{ren2023domain}} & \bf{95.20} & 91.03 & 86.35 & 92.53 & 92.99 & 85.87 & 88.53 & 92.48 & 91.53 & 87.72 & 86.21 & 96.08 & 90.54\\
& {\bf BPL~\cite{wang2023better}} & 94.63 & 90.88 & 87.25 & 92.42 & \underline{93.71} & \underline{86.66} & 88.00 & 92.73 & \underline{92.03} & 88.48 & 86.81 & \underline{96.13} & 90.81\\
& {\bf ADGN~\cite{li2022new}} & 95.03 & 90.67 & 87.39 & 91.87 & 93.52 & 86.34 & 87.98 & 93.50 & 91.54 & 88.02 & \underline{87.55} & 95.28 & 90.72\\
& {\bf ECB-FAM-DG~\cite{ma2021unsupervised}} & 94.39 & 90.94 & \underline{87.40} & \underline{93.04} & 93.00 & 85.86 & 88.36 & \underline{93.99} & 91.84 & \bf{89.30} & 87.48 & 96.07 & \underline{90.97}\\
\cline{2-15}
& {\bf MSADGN (Ours)} & \underline{95.17} & \underline{91.44} & \bf{88.48} & \bf{94.80} & \bf{95.38} & \bf{87.42} & \bf{90.00} & \bf{94.09} & \bf{93.52} & \underline{89.27} & \bf{88.01} & \bf{97.73} & \bf{92.11}\\
\bottomrule
\end{tabular}
}
\end{table*}

\begin{table*}[!t]
\centering
\caption{Classification Accuracy (\%) with State-of-the-Art Methods on CS-HRRSI (The Title in The First Row Denotes ${\mathcal D}_t$ and in The Second Row Denotes ${\mathcal D}_s^1$, and The Best Results Are in Bold and Second-Best Results Are Underlined)}
\label{table: Classification Accuracy on CS-HRRSI}
\centering
\resizebox{\linewidth}{!}{
\begin{tabular}{llccccccccccccc}
\toprule
   \multicolumn{2}{c}{\bf Method} & \multicolumn{3}{c}{\bf N} & \multicolumn{3}{c}{\bf A} & \multicolumn{3}{c}{\bf U} & \multicolumn{3}{c}{\bf W} & {\bf Avg} \\
   \cmidrule(r){3-5} \cmidrule(r){6-8}
   \cmidrule(r){9-11}
   \cmidrule(r){12-14}
    & & {\bf A} & {\bf U} & {\bf W} & {\bf N} & {\bf U} & {\bf W} & {\bf N} & {\bf A} & {\bf W} & {\bf N} & {\bf A} & {\bf U} \\
\midrule
{\bf BL} & {\bf ERM~\cite{he2016deep}} & 85.14 & 64.17 & 73.12 & 94.46 & 67.01 & 86.93 & 84.82 & 75.64 & 65.55 & 95.21 & 98.75 & 71.88 & 80.22\\
\hline
\multirow{6}*{\bf DA} & {\bf DDC~\cite{tzeng2014deep}} & 88.48 & 64.89 & 73.42 & 95.14 & 69.12 & 87.33 & 86.09 & 76.64 & 67.90 & 97.08 & 99.13 & 72.71 & 81.49\\
& {\bf DAN~\cite{long2015learning}} & 89.21 & 66.67 & 73.85 & 94.89 & 70.40 & 87.71 & 86.45 & 76.27 & 65.64 & 96.63 & 99.33 & 73.12 & 81.68\\
& {\bf DANN~\cite{ganin2015unsupervised}} & 89.04 & 76.21 & 73.57 & 95.67 & 80.53 & 87.51 & 85.36 & 80.09 & 67.91 & 96.46 & 99.36 & 79.58 & 84.27\\
&{\bf ECB-FAM~\cite{ma2021unsupervised}} & 89.21 & 73.23 & 74.56 & 94.91 & 78.52 & 87.68 & 87.00 & 79.55 & 66.00 & 96.67 & 99.42 & 81.87 & 84.05\\
& {\bf TLADAN~\cite{zhang2023triple}} & 89.15 & 75.45 & 75.02 & 96.20 & 81.22 & 87.08 & 86.73 & 78.00 & 66.00 & 97.29 & 99.38 & 81.67 & 84.43\\
\hline
\multirow{6}*{\bf DG} & {\bf DSDGN~\cite{liao2020deep}} & 89.87 & \underline{79.93} & \bf{79.00} & 95.84 & 80.05 & 90.13 & 87.18 & 79.55 & 67.55 & 96.04 & 98.75 & 83.75 & 85.64\\
& {\bf DFGN~\cite{ren2023domain}} & \bf{91.61} & 76.07 & 77.24 & 95.67 & 77.99 & 89.47 & 86.73 & 78.73 & 66.55 & \underline{97.71} & 99.58 & 77.92 & 84.61\\
& {\bf BPL~\cite{wang2023better}} & 90.27 & 77.74 & 78.43 & \underline{96.65} & 79.92 & \bf{92.37} & 86.73 & 78.55 & \bf{71.36} & 96.46 & \underline{99.79} & 78.33 & 85.55\\
& {\bf ADGN~\cite{li2022new}} & 90.33 & 77.55 & 77.02 & 96.25 & 79.04 & 89.82 & \underline{87.36} & \underline{80.09} & 70.00 & 97.29 & \bf{99.88} & \bf{86.04} & 85.89\\
& {\bf ECB-FAM-DG~\cite{ma2021unsupervised}} & 90.75 & 75.80 & 77.76 & 95.97 & \underline{83.07} & 92.17 & \bf{88.73} & 79.35 & 70.09 & 97.08 & 99.38 & 84.00 & \underline{86.18}\\
\cline{2-15}
& {\bf MSADGN (Ours)} & \underline{90.93} & \bf{80.68} & \underline{78.49} & \bf{96.69} & \bf{84.87} & \underline{92.19} & 87.01 & \bf{80.64} & \underline{70.82} & \bf{98.01} & 98.92 & \underline{84.58} & \bf{86.99}\\
\bottomrule
\end{tabular}
}
\end{table*}

\subsection{Ablation Study}
\label{sub: Ablation Study}
An ablation study is conducted to demonstrate the effectiveness of 
different components of MSADGN.
We randomly select two cross-scene scenarios on CS-SLCS and CS-HRRSI, respectively, and obtain the results as shown in Table~\ref{table: Ablation Study Results} using the following methods.
\begin{itemize}
    \item[1)] \textit{M1}.
    It refers to ERM, which does not include a domain-related pseudolabeling module, domain-invariant module, or domain-specific module, serving as a BL method.
    
    \item[2)] \textit{M2}.
    It does not take into account domain-related information in the pseudolabeling module, that is, the traditional pseudolabel method is employed in this module.
    \item[3)] \textit{M3}.
    It replaces the dynamic threshold of MSADGN with a traditional fixed threshold.
    Specifically, setting $p=0$ in Eq.~\eqref{equ: tau} yields $\tau_0=0.4$.
    
    \item[4)] \textit{M4}.
    It refers to using the local iteration scheme instead of the global iteration scheme to update ${\mathbf M}_1$ in Eq.~\eqref{equ: M_1}.
    
    \item[5)] \textit{M5}.
    It removes the domain-invariant module to assess the importance of the domain-invariant feature for cross-scene classification.
    
    \item[6)] \textit{M6}.
    It removes the domain-specific module to assess the importance of the domain-specific features for cross-scene classification.
    
    \item[7)] \textit{M7}.
    It refers to the full version of MSADGN.
\end{itemize}

From the experimental results, it can be observed that the performance of M2 to M7 is superior to that of M1.
This is because the domain-invariant or domain-specific features learned by these methods help reduce the distribution difference between the multisource domains and the target domain.
Among them, our proposed M7 achieves the best performance.
This is due to the effective utilization of unlabeled source domains by the domain-related pseudolabeling module, as well as the more comprehensive features learned by the domain-invariant module and domain-specific module.
Therefore, the effectiveness of different components of MSADGN is verified.

\begin{table*}[!t]
\centering
\caption{Ablation Study Results (\%), Where the First Domain Denotes ${\mathcal D}_s^1$}
\label{table: Ablation Study Results}
\centering
\resizebox{\linewidth}{!}{
\begin{tabular}{cccccccccccccc}
\toprule
   {\bf Method} & \multicolumn{4}{c}{\bf Domain-related pseudolabeling module} &{\bf Domain-invariant module} & {\bf Domain-specific module} & \multicolumn{2}{c}{\bf CS-SLCS} & \multicolumn{2}{c}{\bf CS-HRRSI} \\
   \cmidrule(r){2-5} 
   \cmidrule(r){8-9}
   \cmidrule(r){10-11}
    & {\bf Probability score} & {\bf Similarity score} & {\bf Dynamic threshold} & {\bf Global iteration scheme} & & & {\bf LRS$\rightarrow$M} & {\bf SRM$\rightarrow$L} & {\bf AUW$\rightarrow$N} & {\bf WNA$\rightarrow$U}\\
\midrule
{\bf M1} & \rule[0.5ex]{.6em}{.55pt} & \rule[0.5ex]{.6em}{.55pt} & \rule[0.5ex]{.6em}{.55pt} & \rule[0.5ex]{.6em}{.55pt} & \rule[0.5ex]{.6em}{.55pt} & \rule[0.5ex]{.6em}{.55pt} & 87.48 & 78.23 & 85.14 & 65.55\\
{\bf M2} & \checkmark & \rule[0.5ex]{.6em}{.55pt} & \rule[0.5ex]{.6em}{.55pt} & \rule[0.5ex]{.6em}{.55pt} & \checkmark & \checkmark & 91.38 & 87.06 & 87.32 & 68.64\\
{\bf M3} & \checkmark & \checkmark & \rule[0.5ex]{.6em}{.55pt} & \checkmark & \checkmark & \checkmark & 93.21 & 87.30 & 87.58 & 67.64\\
{\bf M4} & \checkmark & \checkmark & \checkmark & \rule[0.5ex]{.6em}{.55pt} & \checkmark & \checkmark & 91.72 & 86.44 & 89.17 & 68.73\\
{\bf M5} & \checkmark & \checkmark & \checkmark & \checkmark & \rule[0.5ex]{.6em}{.55pt} & \checkmark & 91.60 & 85.86 & 88.69 & 65.82\\
{\bf M6} & \checkmark & \checkmark & \checkmark & \checkmark & \checkmark & \rule[0.5ex]{.6em}{.55pt} & 92.43 & 86.05 & 89.77 & 66.09\\
{\bf M7} & \checkmark & \checkmark & \checkmark & \checkmark & \checkmark & \checkmark & 93.52 & 88.01 & 90.93 & 70.82\\
\bottomrule
\end{tabular}
}
\end{table*}

\subsection{Further Empirical Analysis}
\label{sub: Further Empirical Analysis}
In this section, we conduct further empirical analysis on the hyperparameters, network structure, and convergence of MSADGN to validate the superiority of the method.
The benchmarks utilized include CS-SLCS and CS-HRRSI.
Unless specified otherwise, CS-SLCS is illustrated with the scenario of {\bf LRS$\rightarrow$M}, and CS-HRRSI is demonstrated with the scenario of {\bf AUW$\rightarrow$N}, where the first domain denotes the labeled source domain.

\subsubsection{Hyperparameters of The Domain-Related pseudolabeling Module}
\label{subsubsec: Hyperparameters}
Here, different values of $\alpha$ in Eq.~\eqref{equ: Phi} are set to explore the influence of the proportion between $\phi({\mathbf x}_i^u)$ and $\psi({\mathbf x}_i^u)$ on the performance of MSADGN.
Specifically, we set $\alpha=[0.0, 1.0]$ with a step size of 0.1 and conduct experiments on the benchmarks.
The experimental results are shown in Fig.~\ref{fig: further analysis}\subref{subfig: parameter}, from which the following conclusions can be drawn.
When $\alpha$ is larger, MSADGN performs poorly on the benchmarks.
This could be attributed to the dominance of the traditional $\phi({\mathbf x}_i^u)$, especially when $\alpha=1.0$, resulting in the worst performance due to the exclusive presence of $\phi({\mathbf x}_i^u)$.
Conversely, when $\alpha$ is smaller, MSADGN demonstrates better performance on the benchmarks, with the optimal performance observed at $\alpha=0.2$.
This is owing to the incorporation of $\psi({\mathbf x}_i^u)$.
Therefore, as mentioned earlier, we set $\alpha=0.2$ in all experiments.

\subsubsection{Discriminators of The Domain-Invariant Module}
\label{subsubsec: Discriminators}
In the domain-invariant module, the $K(K-1)/2$ binary discriminators are replaced by a single $K$-class discriminator to investigate the influence of the number of discriminators on MSADGN.
It can be seen from Fig.~\ref{fig: further analysis}\subref{subfig: discriminator} that the performance of the former surpasses that of the latter on the benchmarks.
One possible reason for this is that the latter simultaneously extracts the domain-invariant feature from $K$ multisource domains, which may increase confusion between categories and make it difficult for the model to accurately generalize to the unseen target domain.

\subsubsection{Weights of The Domain-Specific Module}
\label{subsubsec: Weights}
The domain-related similarity weights of a sample randomly selected from the unseen target domain of the benchmarks are shown in Fig.~\ref{fig: further analysis}\subref{subfig: weight}.
As mentioned earlier, the weights imply the similarity between the target domain and the multisource domains.
Therefore, MSADGN effectively learns the similarity between the target domain and each source domain, so that the prediction results can benefit from the domain-specific features learned from the multisource domains.

\subsubsection{Convergence of The Proposed MSADGN}
\label{subsubsec: Convergence}
\begin{figure*}[!t]
\centering
\subfloat[]{\includegraphics[width=1.6in]{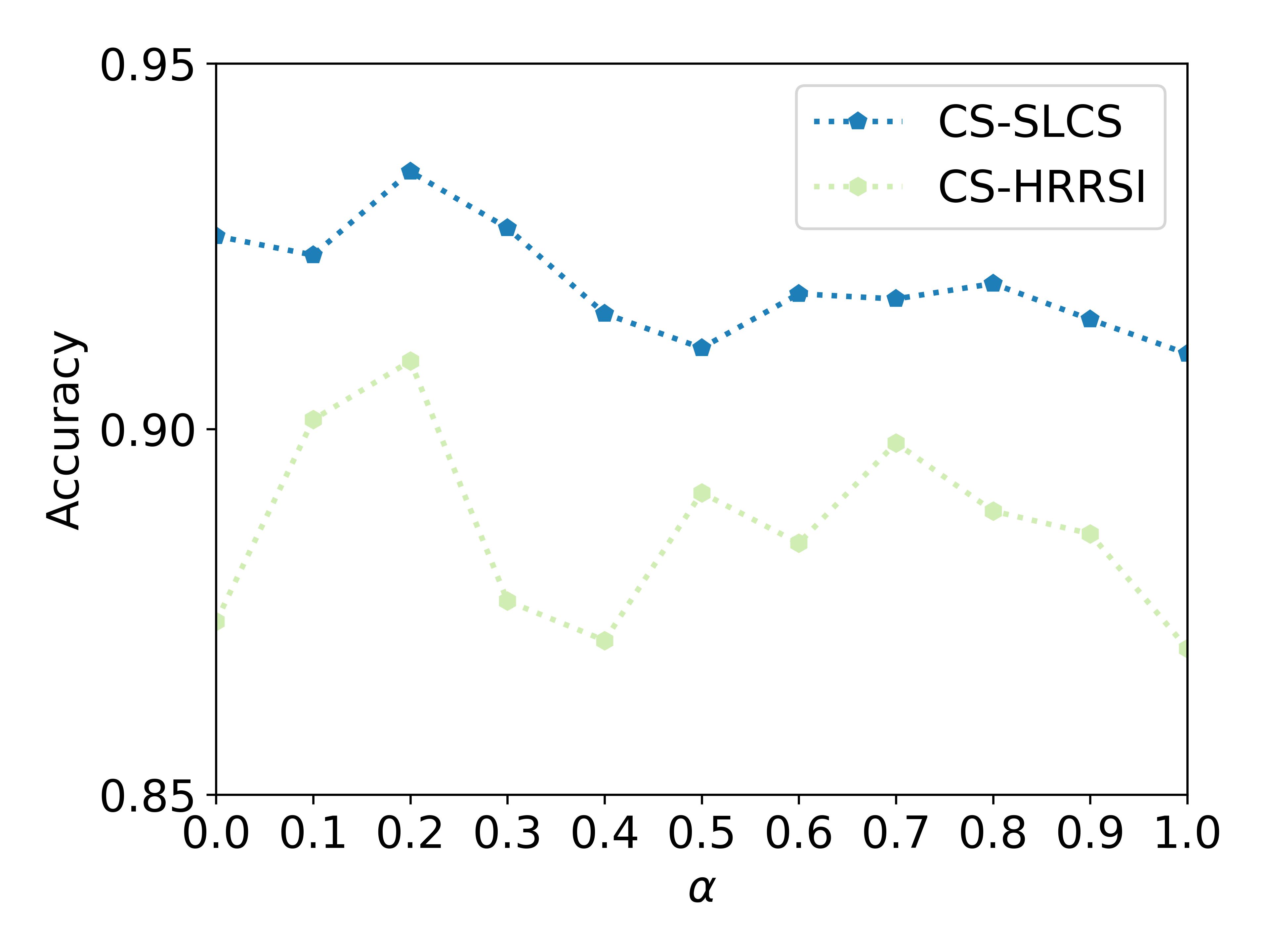}
\label{subfig: parameter}}
\hspace{0pt}
\subfloat[]{\includegraphics[width=1.6in]{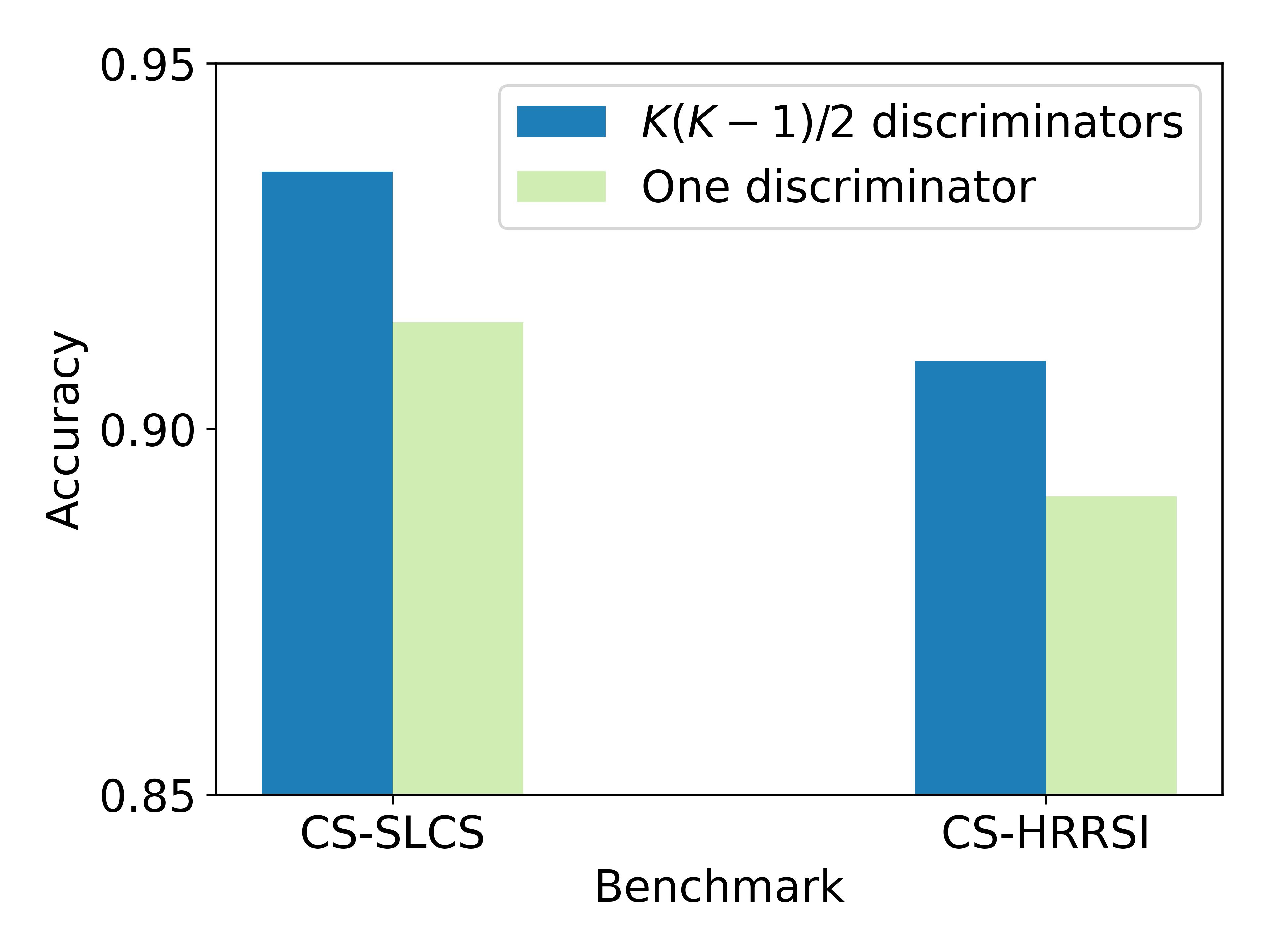}
\label{subfig: discriminator}}
\hspace{0pt}
\subfloat[]{\includegraphics[width=1.6in]{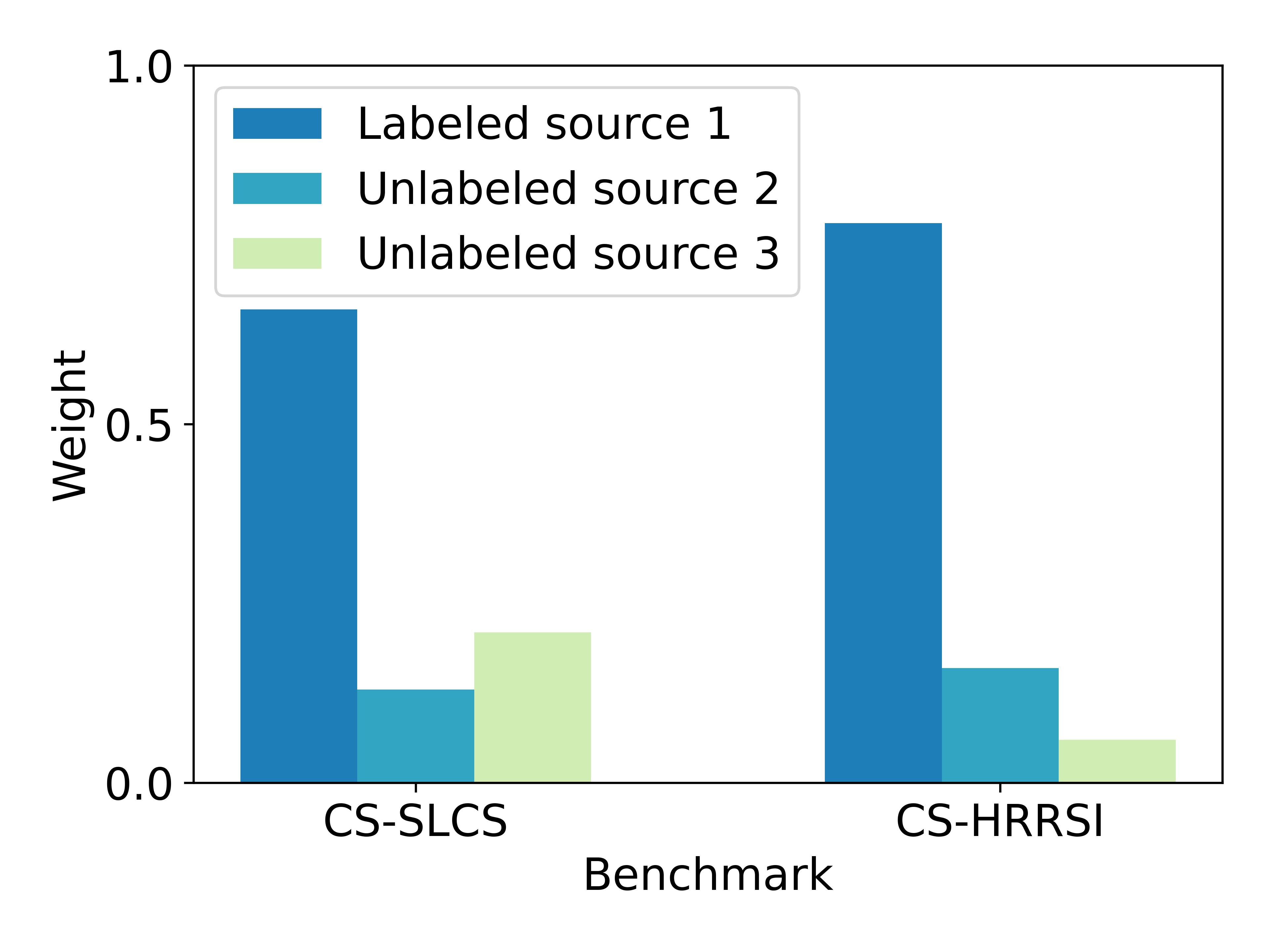}
\label{subfig: weight}}
\hspace{0pt}
\subfloat[]{\includegraphics[width=1.6in]{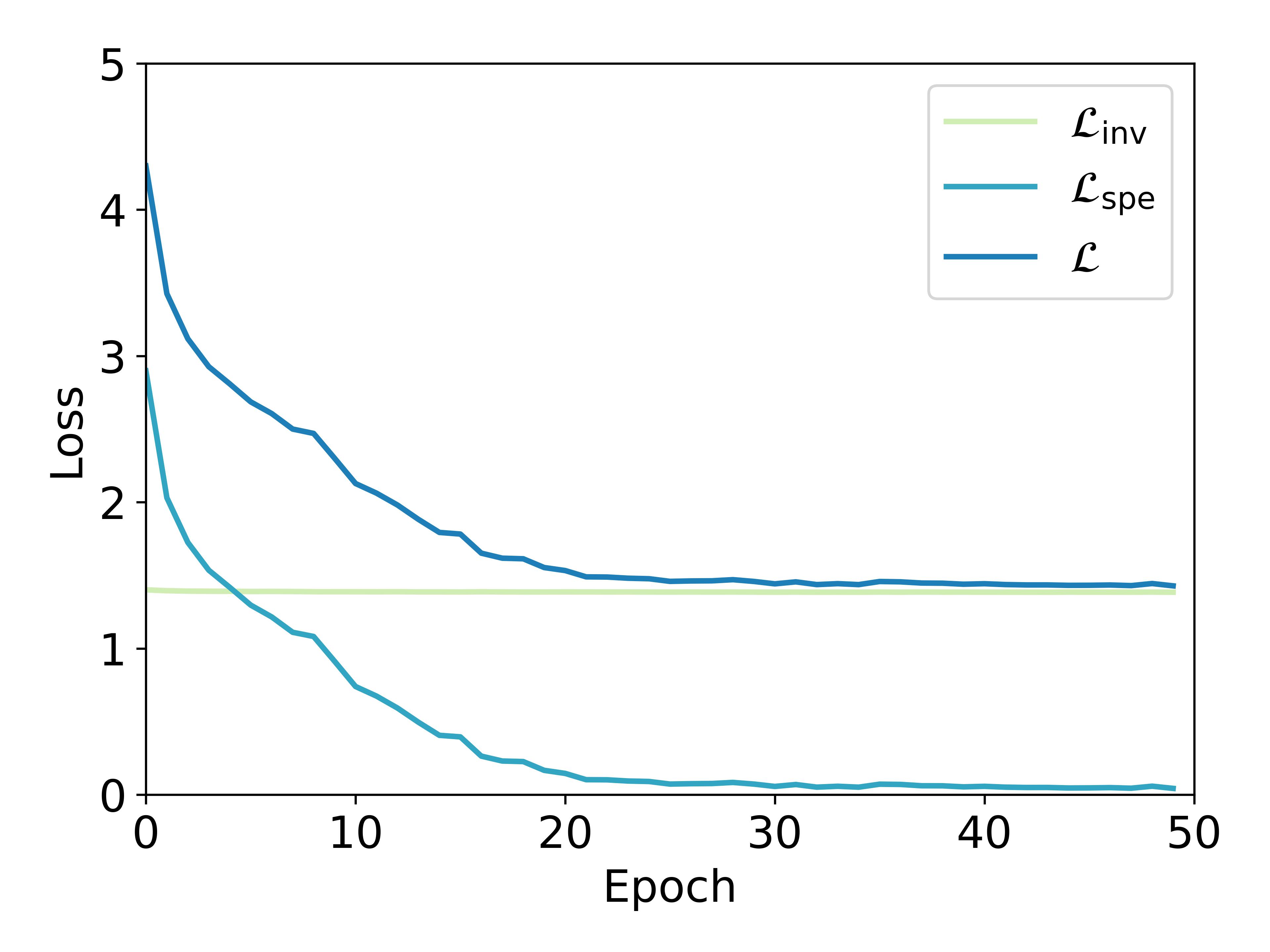}
\label{subfig: loss}}
\caption{Further empirical analysis of MSADGN. (a) Sensitivity analysis of $\alpha$ in the domain-related pseudolabeling module on benchmarks. (b) Analysis of the number of discriminators in the domain-invariant module. (c) Weights of a sample randomly selected from the unseen target domain of benchmarks. (d) Loss curve of MSADGN on CS-HRRSI.}
\label{fig: further analysis}
\end{figure*}

we further empirically analyze the convergence of the proposed MSADGN. Fig.~\ref{fig: further analysis}\subref{subfig: loss} plots the loss curve of MSADGN on CS-HRRSI.
Similar observations are noted for CS-SLCS.
The experimental results show that the training of MSADGN is stable and converges rapidly.

\subsubsection{Categorywise Accuracy}
\label{subsubsec: Categorywise Accuracy}
\begin{figure*}
\centering
\subfloat[]{
\begin{minipage}[t]{0.32\textwidth}
\centering
\includegraphics[width=2.1in]{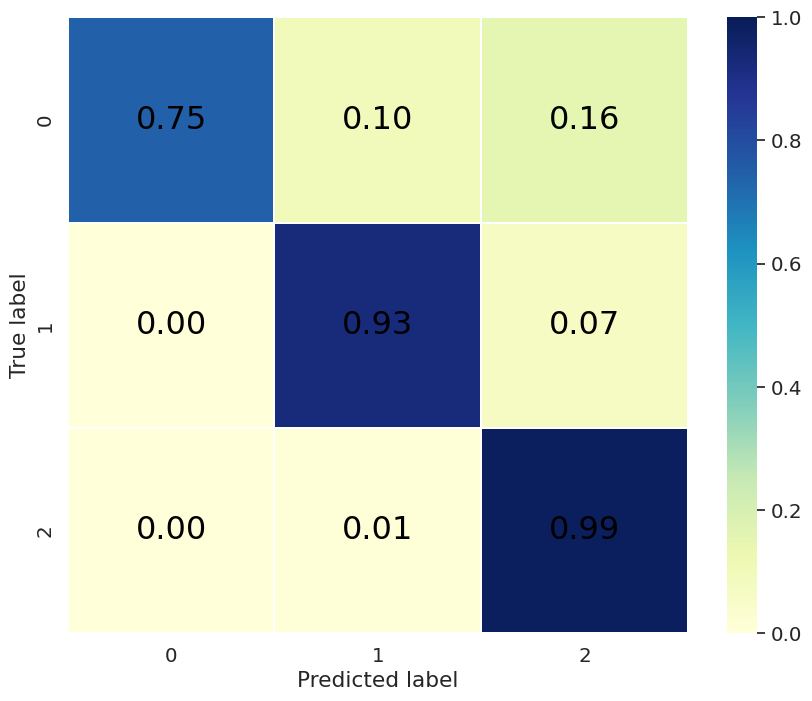}\\
\vspace{0.01cm}
\includegraphics[width=2.1in]{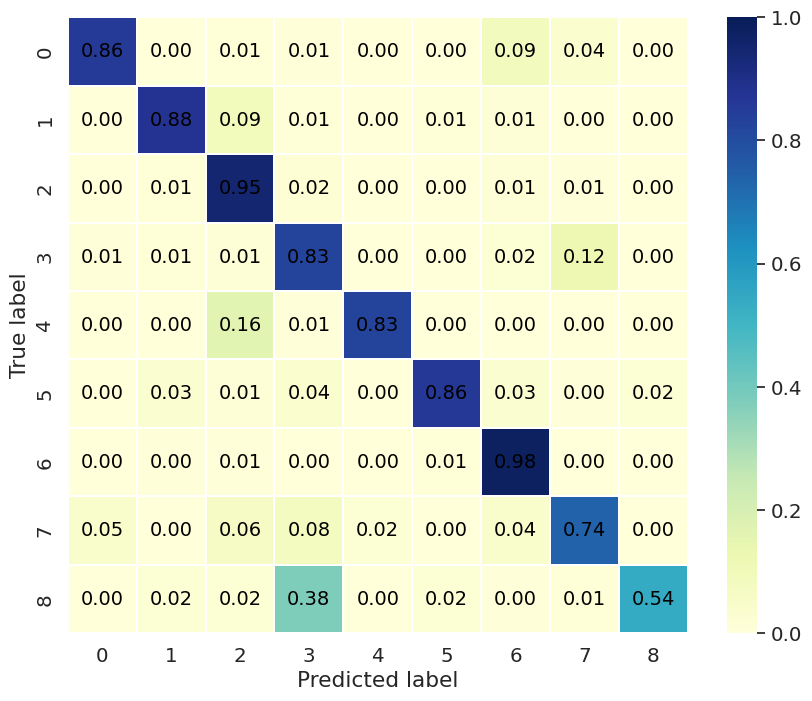}\\
\label{subfig: confusion_ERM}
\end{minipage}}
\hspace{0pt}
\subfloat[]{
\begin{minipage}[t]{0.32\textwidth}
\centering
\includegraphics[width=2.1in]{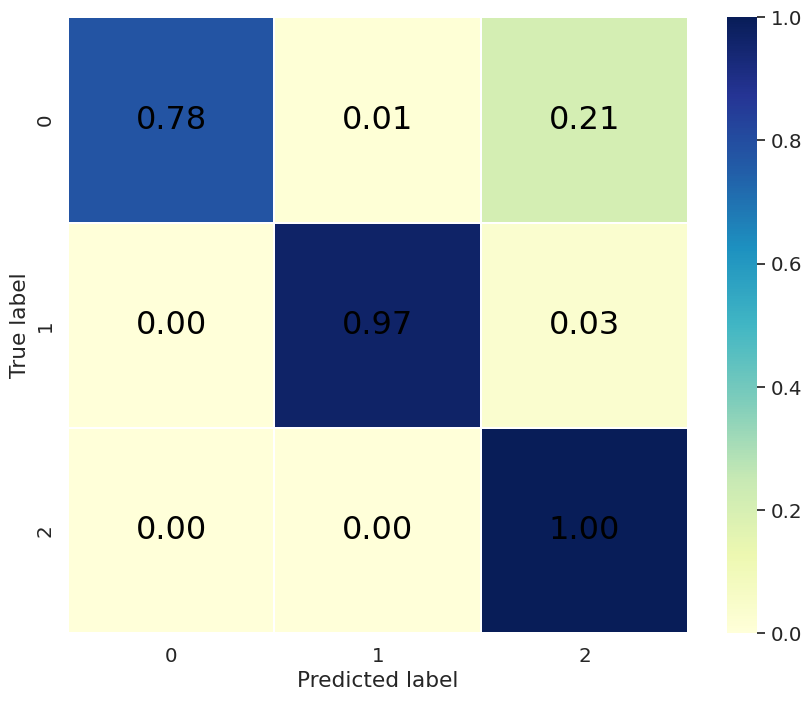}\\
\vspace{0.01cm}
\includegraphics[width=2.1in]{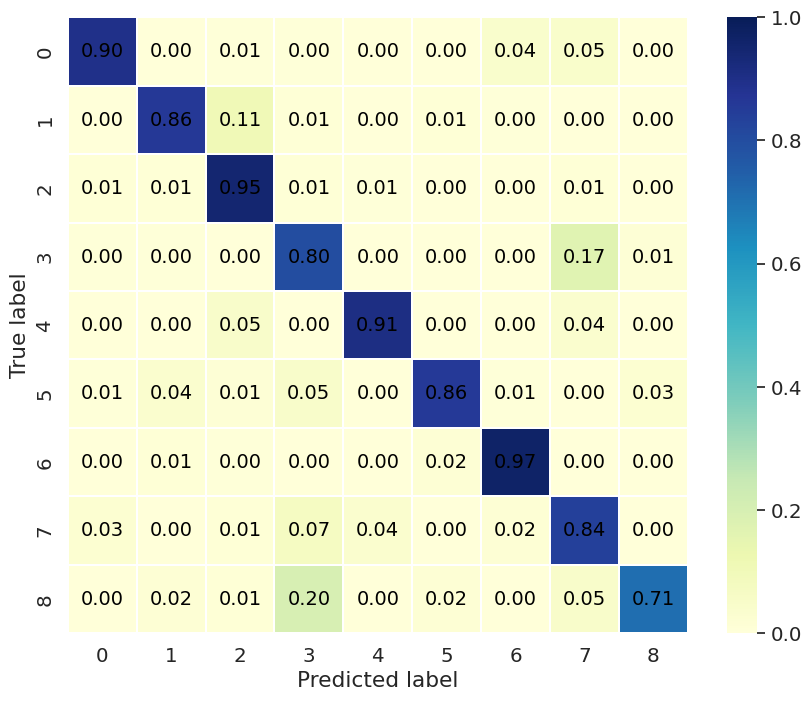}\\
\label{subfig: confusion_DSDGN}
\end{minipage}}
\hspace{0pt}
\subfloat[]{
\begin{minipage}[t]{0.32\textwidth}
\centering
\includegraphics[width=2.1in]{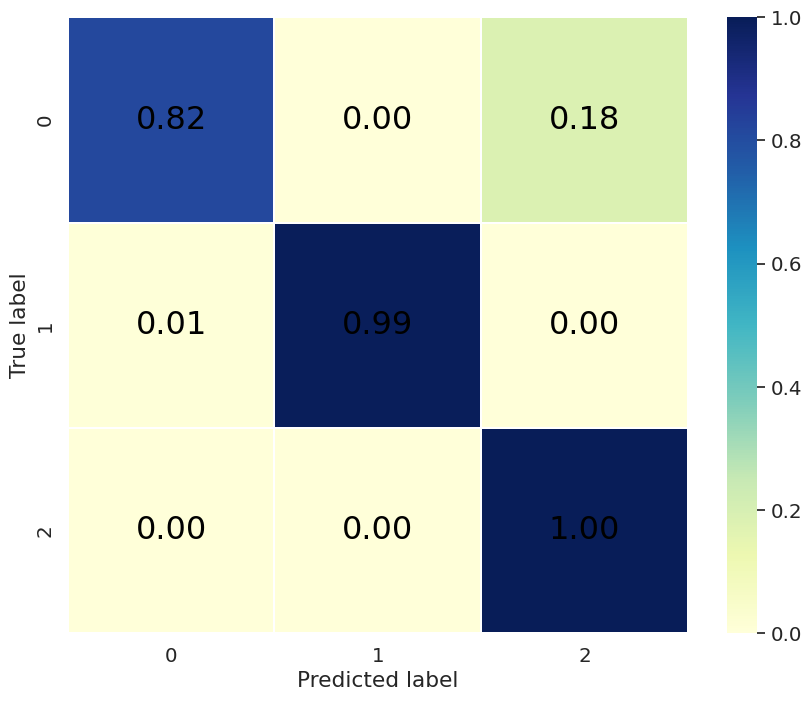}\\
\vspace{0.01cm}
\includegraphics[width=2.1in]{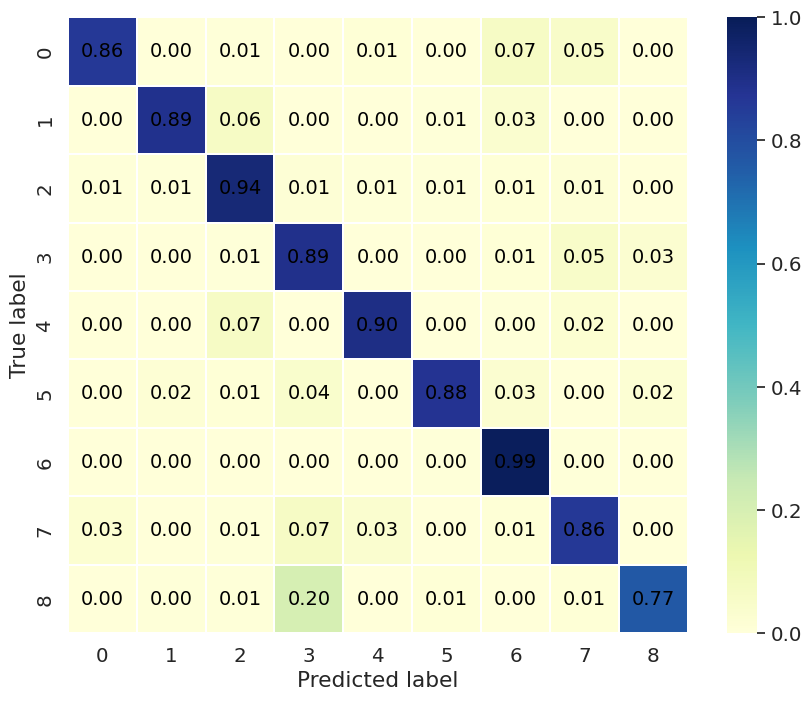}\\
\label{subfig: confusion_MSADGN}
\end{minipage}}
\caption{Confusion matrices on (top) CS-SLCS and (bottom) CS-HRRSI. (a) ERM. (b) DSDGN. (c) MSADGN.}
\label{fig: Confusion Matrices}
\end{figure*}

It should be emphasized that MSADGN is an improved version of DSDGN.
To demonstrate the superiority of our method over DSDGN, we utilize the confusion matrix for categorywise accuracy analysis on the benchmarks.
For both CS-SLCS and CS-HRRSI, consistent conclusions can be drawn from Fig.~\ref{fig: Confusion Matrices}, indicating that the performance of both DSDGN and MSADGN significantly surpasses that of ERM.
This is attributed to their capability for DG.
Furthermore, the performance of MSADGN surpasses that of DSDGN, owing to its superior domain-related pseudolabeling module and the exclusive domain-specific module.
For example, ERM exhibits poor classification accuracy (75\%) for category 0 on CS-SLCS, whereas both DSDGN and MSADGN show significant improvements in classification accuracies of 78\% and 82\% for category 0, respectively.
Similarly, ERM exhibits poor classification accuracy (74\%) for category 7 and (54\%) for category 8 on CS-HRRSI. In contrast, both DSDGN and MSADGN show significant improvements in classification accuracies of 84\% and 86\% for category 7, and 71\% and 77\% for category 8, respectively.
Therefore, the superiority of MSADGN in categorywise accuracy is validated.

\subsubsection{Feature Visualization}
\label{subsubsec: Feature Visualization}
\begin{figure*}[!t]
\centering
\subfloat[\qquad\qquad\qquad]{\includegraphics[width=2.1in]{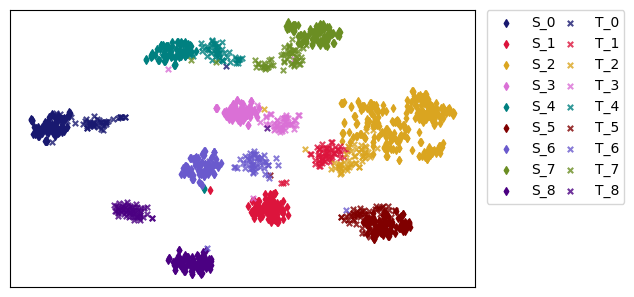}
\label{subfig: tsne_ERM}}
\hspace{0pt}
\subfloat[\qquad\qquad\qquad]
{\includegraphics[width=2.1in]{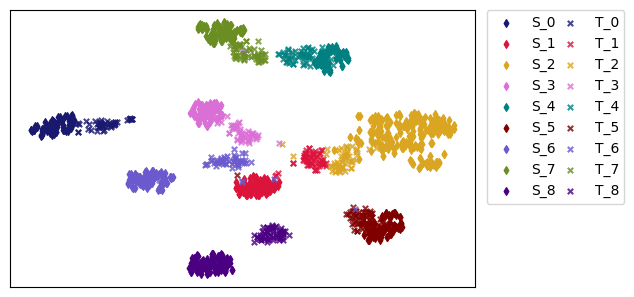}
\label{subfig: tsne_DAN}}
\hspace{0pt}
\subfloat[\qquad\qquad\qquad]
{\includegraphics[width=2.1in]{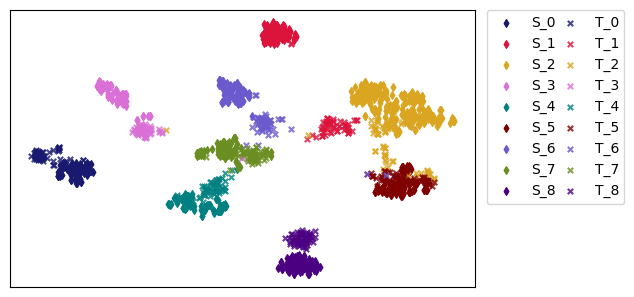}
\label{subfig: tsne_DANN}}\\
\vspace{0.01cm}
\subfloat[\qquad\qquad\qquad]
{\includegraphics[width=2.1in]{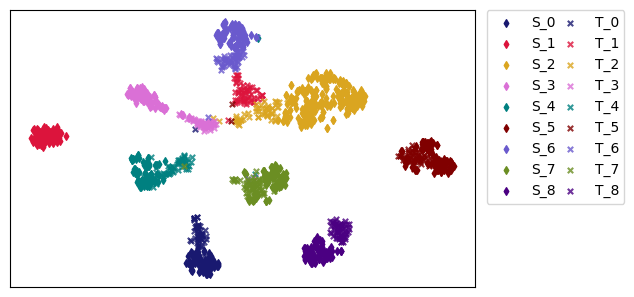}
\label{subfig: tsne_TLADAN}}
\hspace{0pt}
\subfloat[\qquad\qquad\qquad]
{\includegraphics[width=2.1in]{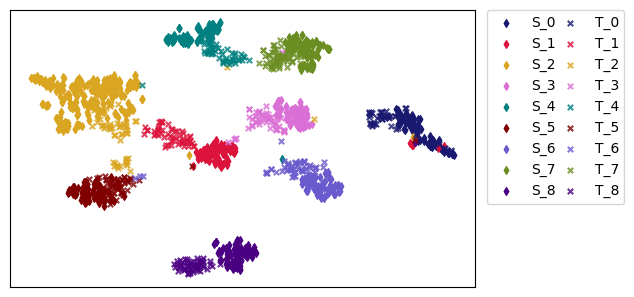}
\label{subfig: tsne_DSDGN}}
\hspace{0pt}
\subfloat[\qquad\qquad\qquad]
{\includegraphics[width=2.1in]{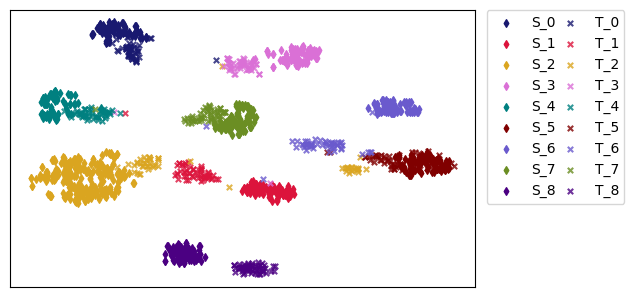}
\label{subfig: tsne_MSADGN}}
\caption{$\bm t$-SNE feature visualization on {\bf UNA$\rightarrow$W}, where ``S" and ``T" in the legend denote ${\mathcal D}_s^1$ and ${\mathcal D}_t$. (a) ERM. (b) DAN. (c) DANN. (d) TLADAN. (e) DSDGN. (f) MSADGN.}
\label{fig: tsne}
\end{figure*}

Finally, the $\bm t$-distributed stochastic neighbor embedding~($\bm t$-SNE) \cite{van2008visualizing} is utilized to visualize the features learned by $F_{\text{shared}}$ with various methods.
To more intuitively demonstrate the superiority of DG over DA and MSADGN over other methods, we conduct experiments on a randomly selected {\bf UNA$\rightarrow$W} task with large differences in classification accuracy, where $\bf U$ denotes the labeled source domain.
Specifically, six methods are selected, including one ERM as BL, three DA-based DG methods (DAN without adversarial training, DANN with adversarial training, and TLADAN with excellent DA performance proposed in our previous work \cite{zhang2023triple}), and two pure DG methods (DSDGN and the proposed MSADGN).
The experimental results are shown in Fig.~\ref{fig: tsne}, from which the following conclusions can be drawn.
\begin{itemize}
    \item[1)] See Fig.~\ref{fig: tsne}\subref{subfig: tsne_ERM}, the features learned by ERM exhibit significant confusion among categories, particularly between categories $\text{T}_1$ and $\text{T}_2$.
    This is attributed to the domain bias between the source and target domains, and ERM does not incorporate DG techniques.

    \item[2)] See Figs.~\ref{fig: tsne}\subref{subfig: tsne_DAN}, \ref{fig: tsne}\subref{subfig: tsne_DANN} and \ref{fig: tsne}\subref{subfig: tsne_TLADAN}, after applying DA-based DG methods, the feature alignment effect is improved compared to ERM.
    Notably, the improvement effect of DAN is not obvious, while DANN and TLADAN exhibit more significant improvements, with almost all features of the same categories gathered together.
    This is due to the adversarial training used by the latter to learn more effective features.

    \item[3)] See Figs.~\ref{fig: tsne}\subref{subfig: tsne_DSDGN} and \ref{fig: tsne}\subref{subfig: tsne_MSADGN}, after using pure DG methods DSDGN and MSADGN, the intraclass aggregation and interclass separation between features are more obvious than those of the DA-based DG methods.
    Although the intraclass aggregation ability of MSADGN is slightly inferior to that of DSDGN (e.g., in categories $\text{T}_1$, $\text{T}_6$, and $\text{T}_8$), MSADGN demonstrates better interclass separation ability (e.g., between categories $\text{T}_1$ and $\text{T}_3$, and between categories $\text{T}_4$ and $\text{T}_7$).
    This makes the nine feature clusters from the target domain learned by MSADGN more distinguishable.
    Therefore, MSADGN performs the best among all methods, benefiting from the more comprehensive features learned by the domain-specific module.
\end{itemize}

\section{Conclusion}
\label{sec: Conclusion}
This study proposed a novel MSADGN for cross-scene sea\textendash land clutter classification.
We considered a more practical yet complex SSDG scenario in which only one labeled source domain and multiple unlabeled source domains were available.
Specifically, an improved pseudolabel method called domain-related pseudolabel was proposed to generate reliable pseudolabels.
Furthermore, the domain-invariant features learned by a multidiscriminator and the domain-specific features learned by a multiclassifier enabled MSADGN to exhibit excellent transferability and discriminability in the unseen target domain.
Our method outperformed state-of-the-art DG methods in numerous cross-scene classification tasks.

Future research involves more complex sea\textendash land clutter classification.
For example, if only one labeled source domain is available, our model will be greatly limited.
Therefore, it is necessary to expand the investigation into single-source DG.

\bibliographystyle{IEEEtran}
\bibliography{references}

\end{document}